\documentclass[letterpaper, 10 pt, journal, twoside]{IEEEtran}  

\usepackage{amsmath}
\usepackage{amssymb}
\usepackage{mathtools}
\usepackage{graphicx}
\usepackage{float}
\usepackage{array}
\usepackage{tikz}
\usepackage{listings}
\usepackage{hyperref}
\usepackage{graphicx}
\usepackage{cite}
\usepackage{dsfont}
\usepackage{mathtools,etoolbox}
\usepackage[ruled, linesnumbered]{algorithm2e}
\usepackage[none]{hyphenat}
\usepackage[utf8]{inputenc}
\usepackage[english]{babel}

\usepackage{amsthm}
\usepackage{xfrac}
\usepackage{nicefrac}
\usepackage{booktabs}
\usepackage{flushend}
\usepackage[switch, pagewise]{lineno}
\usepackage{mathrsfs}
\DeclareMathAlphabet{\mathpzc}{OT1}{pzc}{m}{it}
\usepackage{multirow}
\usepackage{multicol}
\usepackage{bbm}
\usepackage{soul}
\usepackage{xfrac}
\usepackage{balance}
\usepackage{color, colortbl}
\usepackage{epstopdf}
\usepackage{footnote}
\makesavenoteenv{tabular}
\makesavenoteenv{table}
\usepackage{fontawesome} 
\definecolor{LightCyan}{rgb}{0.88,1,1}
\definecolor{LightOrange}{rgb}{1,0.7,0}

\DeclareMathOperator*{\argmin}{argmin} 

\hypersetup{colorlinks=true,urlcolor=blue,citecolor=red}

\title{\LARGE \bf PRGFlow: Benchmarking SWAP-Aware Unified Deep Visual Inertial Odometry}
\author{Nitin J. Sanket, Chahat Deep Singh, Cornelia Ferm{\"u}ller, Yiannis Aloimonos 
\thanks{\textit{Corresponding author: Nitin J. Sanket.} All the authors are with Perception and Robotics Group, University of Maryland Institute for Advanced Computer Studies, University of Maryland, College Park.}
} 

\begin{document}




\maketitle

\begin{abstract}
Odometry on aerial robots has to be of  low latency and  high robustness whilst also respecting the Size, Weight, Area and Power (SWAP) constraints as demanded by the size of the robot. A combination of visual sensors coupled with Inertial Measurement Units (IMUs) has proven to be the best combination to obtain robust and low latency odometry on resource-constrained aerial robots. Recently, deep learning approaches for Visual Inertial fusion have gained momentum due to their high accuracy and robustness. However, the remarkable advantages of these techniques are their inherent scalability (adaptation to different sized aerial robots) and unification (same method works on different sized aerial robots) by utilizing compression methods and hardware acceleration, which have been lacking from previous approaches.

To this end, we present a deep learning approach for visual translation estimation and loosely fuse it with an Inertial sensor for full 6DoF odometry estimation. We also present a detailed benchmark comparing different architectures, loss functions and compression methods to enable scalability. We evaluate our network on the MSCOCO dataset and evaluate the VI fusion on multiple real-flight trajectories.
\end{abstract}



\section{Introduction}
A fundamental competence of  aerial robots \cite{gapflyt} is to estimate ego-motion or odometry before any control strategy is employed \cite{Exploration, SearchAndRescue, fermuller93, mcguire2019minimal}. Different sensor combinations have been used previously to aid the odometry estimation with LIDAR based approaches topping the accuracy charts \cite{loam, vloam}. However such approaches cannot be used on smaller aerial robots due to their size, weight and power constraints. Such small aerial robots are generally preferred due to safety, agility and usability as swarms \cite{morbidi2011estimation, weinstein2018vio, shishika2019mosquito}. In the last decade, imaging sensors have struck the right balance considering  accuracy and general  sensor utility \cite{ovc}. However, visual data is dense and requires a lot of  computation, which creates challenges for low-latency applications. To this end, sensor fusion experts proposed to use IMUs along with imaging sensors, because IMUs are lightweight and are generally available on aerial robots \cite{MSCKF}. Also, employing IMUs with even a monocular camera enables the estimation of metric depth which can be useful for many applications.

  \begin{figure*}[t!]
     \centering
     \includegraphics[width=\textwidth]{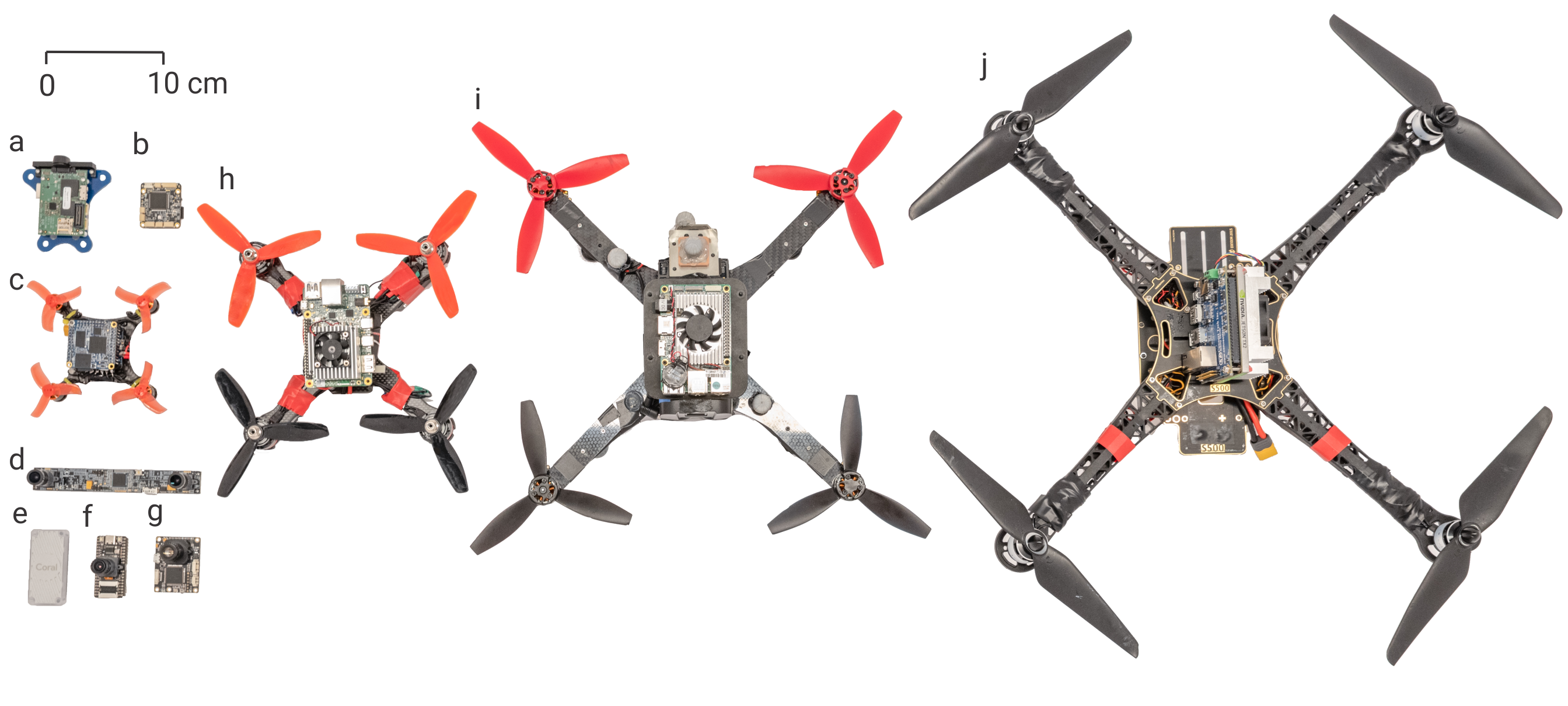}
     \caption{Size comparison of various components used on quadrotors. (a) Snapdragon Flight, (b) PixFalcon, (c) 120 mm quadrotor platform with NanoPi Neo Core 2, (d) MYNT EYE stereo camera, (e) Google Coral USB accelerator, (f) Sipeed Maix Bit, (g) PX4Flow, (h) 210 mm quadrotor platform with Coral Dev board, (i) 360 mm quadrotor platform with Intel$^\text{\textregistered}$ Up board, (j) 500 mm quadrotor platform with NVIDIA$^\text{\textregistered}$ Jetson$^\text{TM}$ TX2. Note that all components shown are to relative scale. \textit{All the images in this paper are best viewed in color.}}
     \label{fig:DiffQuadrotors}
 \end{figure*}

In the last decade, several VIO algorithms have been used in commercial products and also many algorithms  have been made open-source by the research community. However, there is no trivial way of downscaling these algorithms for  smaller aerial robots \cite{evdodgenet}.

In the last five years, deep learning based approaches for visual and visual inertial odometry estimation have gained momentum. We classify as such algorithms all approaches which learn to predict odometry in an end-to-end fashion using one of the aforementioned sensors or which use deep learning as a part of the odometry estimation. The networks  used in these approaches can be compressed  to smaller size with generally a linear drop in accuracy to cater to  SWAP constraints. The critical issue with deep networks for odometry estimation is that to have the  same accuracy as classical approaches they are generally computationally heavy
leading to larger latencies. However, leveraging hardware acceleration and better parallelizable architectures can mitigate this problem.

In this work, we present a method for visual inertial odometry estimation targeted towards a down-facing/up-facing camera coupled to an altimeter source such as a barometer (outdoor) or SONAR or single beam LIDAR (indoor). Our approach uses deep learning to obtain translation - shift and zoom-in/out and/or yaw. The inputs to the network are rotation compensated using Inertial estimates of attitude. Finally, we use the altimeter to scale the shifts to real-world velocities similar to \cite{px4flow}. We also benchmark different combinations of our approach and answer the following questions: \textit{How many warping blocks to use? What network architecture to use? Which loss function to use? What is the best way to compress? Which common hardware is the best for a certain-sized aerial robot?}

\subsection{Problem Definition and Contributions}
A quadrotor is equipped with a down/up-facing camera coupled to an altimeter and an IMU. The aim is to estimate ego-motion or odometry combining all sources of information. The presented approach has to be \textit{scalable} and \textit{unified} so that the same method can be used on different sized aerial robots catering to different SWAP constraints (Fig. \ref{fig:DiffQuadrotors} shows examples of different sized quadrotors with different components which can be used on quadrotors).

A summary of our contributions is given below:
\begin{itemize}
    \item A deep learning approach to estimate odometry using visual, inertial and altimeter data
    \item A comprehensive benchmark of different network architectures, hardware architectures and loss functions
    \item Real-flight experiments demonstrating robustness of the presented approach
    \item Notes to practitioners whenever applicable
\end{itemize}

 \begin{figure*}[t!]
     \centering
     \includegraphics[width=\textwidth]{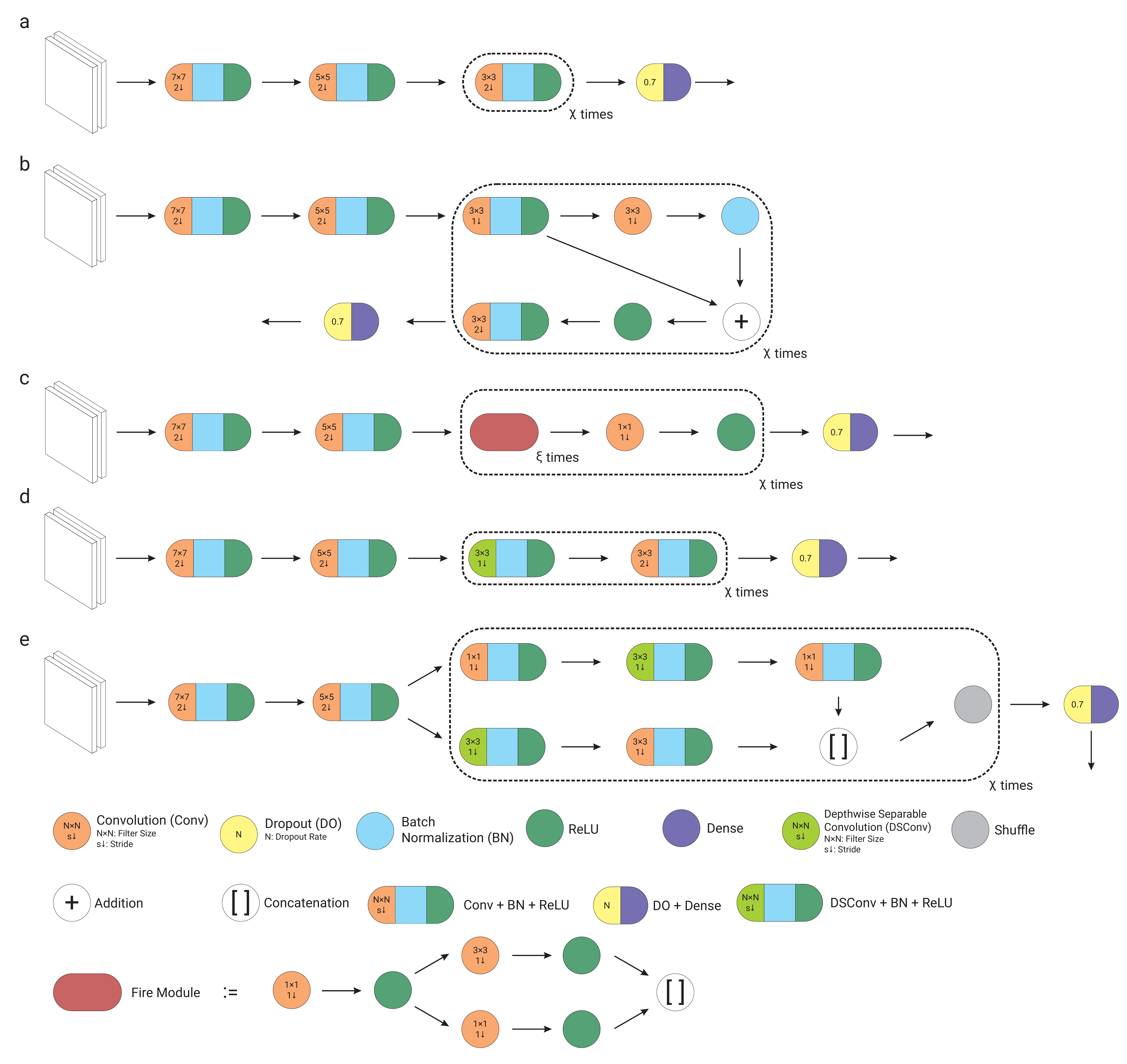}
     \caption{Different network architectures. (a) VanillaNet, (b) ResNet, (c) SqueezeNet, (d) MobileNet and (e) ShuffleNet. ($\chi$ and $\xi$ are hyperparameters). Each architecture block is repeated per warp parameter prediction. \textit{This image is best viewed on the computer screen at a zoom of 200\%}.}
     \label{fig:NetworkArchitectures}
 \end{figure*}

\subsection{Related Work}
There has been extensive progress in the field of Visual or Visual-Inertial Odometry (VI or VIO) using classical approaches, but adapting them to  deep learning is still in a nascent stage. We categorize related work into the following three categories: VI/VIO using classical methods (non-deep learning), deep learning based VI/VIO, and deep learning and odometry benchmarks. Also, note that we do not consider Simultaneous Localization And Mapping (SLAM) approaches \cite{SLAMSurvey} such as ORB-SLAM \cite{orbslam}, LSD-SLAM \cite{lsdslam}, LOAM \cite{loam}, V-LOAM \cite{vloam} and probabilistic object centric slam \cite{bowman2017probabilistic}. We also exclude LIDAR and SONAR based odometry approaches from our discussion.

\subsubsection{VI/VIO using classical methods}
Following are the state-of-the-art approaches in chronological order.
\begin{itemize}
\item MSCKF \cite{MSCKF} proposed an Extended Kalman Filter (EKF) for visual inertial odometry.
\item OKVIS \cite{OKVIS} proposed a stereo-keyframe based sliding window estimator to reduce landmark re-projection errors.
\item ROVIO \cite{rovio} also uses an EKF but included tracking 3D landmarks along with tracking of image patches.
\item DSO \cite{DSO} uses a direct approach using a photometric model coupled with  a geometric model to achieve the best compromise of speed and accuracy.
\item VINS-Mono \cite{VINSMono} introduced a non-linear optimization based sliding window estimator with pre-integrated IMU factors.
\item Salient-DSO \cite{salientdso} builds upon DSO to add visual saliency using deep learning for feature extraction. However, the optimization or regression of poses is performed classically.
\end{itemize}

\subsubsection{Deep learning based VI/VIO}
\begin{itemize}
\item PoseNet \cite{posenet} uses a deep network to re-localize a camera in a pre-trained scene which brought the robustness and ease of use of deep networks for camera pose regression into limelight. Better loss functions for the same function were presented in \cite{posenet2}.
\item SfMLearner \cite{SfMLearner} took this one step further to regress camera poses and depth simultaneously from a sequence of video frames in a completely self-supervised (unsupervised) manner using geometric constraints.
\item GeoNet \cite{GeoNet} built upon SfMLearner to add additional geometric constraints and proposed a new training method along with a novel network architecture to predict pose, depth and optical flow in a completely self-supervised (unsupervised) manner.
\item D3VO \cite{D3VO} tightly incorporates the predicted depth, pose and uncertainty into a direct visual odometry method to boost both the front-end tracking as well as the back-end non-linear optimization.
\item VINet \cite{VINet} proposed a supervised method to estimate odometry from a CNN + LSTM combination using both visual and inertial data. This approach, however, does not present results about its generalizability to novel scenes.
\item DeepVIO \cite{DeepVIO} presents an approach to tightly integrate visual and inertial features using a CNN + LSTM to estimate odometry. This method also  does not present results about its generalizability to novel scenes.
\end{itemize}

\subsubsection{Deep learning and Odometry benchmarks}
A myriad of datasets such as KITTI \cite{KITTI}, EuRoC \cite{EUROC}, TUMMonoVO \cite{TUMMonoVO}, and PennCOSYVIO \cite{Penncosyvio} have been proposed to evaluate the performance of VI/VIO approaches, but they do not contain enough images to train a neural network to  generalize to novel datasets.

Though there exist several benchmarks for either classical VO/VIO approaches \cite{delmerico2018benchmark} and for deep learning for classification/regression tasks \cite{DLbenchmark1, DLbenchmark2, scheper2020evolution}, there is a big void in benchmarks for deep learning based VO/VIO approaches, which is the focus of this work.

\section{Pseudo-Similarity Estimation Using PRGFlow}

Let us mathematically formulate our problem statement. Let $\mathcal{I}_t$ and $\mathcal{I}_{t+1}$ be the image frames captured at times $t$ and $t+1$, respectively. Now, the transformation between the image frames can be expressed as $\mathbf{x}_{t+1} = \mathbf{H}_t^{t+1}\mathbf{x}_t$, where $\mathbf{x}_{t+1}, \mathbf{x}_{t}$ represent the homogeneous point correspondences in the two image frames and $\mathbf{H}_t^{t+1}$ is the  non-singular $3 \times 3$ transformation matrix between the two frames.

While in general the transformation between views is a non-linear function of the 3D rotation matrix $\mathbf{R}_t^{t+1}$ and 3D translation vector $\mathbf{T}_t^{t+1}$, for certain scene structures,
the transformation
can simplify to a linear function. One such case is when the real world area is planar or can be approximated to be a plane, or the focal length is large. This scenario is also called a homography with $\mathbf{H}_t^{t+1}$ referred to as the \textit{Homography matrix}.


From $\mathbf{H}_t^{t+1}$ we can recover a finite number of $\{ \mathbf{R}_t^{t+1}, \mathbf{T}_t^{t+1}\}$ solutions. Furthermore, $\mathbf{H}_t^{t+1}$ may also be  decomposed into simpler transformations such as in-plane rotation or yaw, zoom-in/out or scale, translation and out-of-plane rotations or pitch + roll. It is difficult to  accurately  derive $\{ \mathbf{R}_t^{t+1} and  \mathbf{T}_t^{t+1}\}$  from $\mathbf{H}_t^{t+1}$, and the errors in the solutions of the two components are  coupled, i.e., an error in the translation estimate would induce a complementary error in the  rotation estimate; this is highly undesirable.


Complementary sensors help mitigate this problem, and IMU is such a sensor which is present on quadrotors and can provide accurate angle measurements within a small interval. Thus,  the problem of estimating cheap ego-motion reduces to finding $\mathbf{T}_t^{t+1}$ (2D translation and zoom-in/out). Further, one can also obtain zoom-in/out using the altimeter on-board, but such an approach is noisy in a small interval. Hence we will estimate the 2D translation and zoom-in/out transformation which we refer to as ``Pseudo-similarity" since it is one degree of freedom less than the similarity transformation (2D translation, zoom-in/out and yaw). We also call our network which estimates pseudo-similarity  ``PRGFlow''. Note that, one can easily use our work to also estimate yaw without any added effort by changing the warping function. Mathematically, the pseudo-similarity transformation is given in Eq. \ref{eq:PS}, where $W$ and $H$ depict the image width and height, respectively.


 \begin{equation} \mathbf{x}_{t+1} =  \begin{bmatrix} \frac{W}{2} & 0 & \frac{W}{2} \\ 0 & \frac{H}{2} & \frac{H}{2}\\ 0 & 0 & 1  \end{bmatrix}\begin{bmatrix} 1 + s & 0 & t_x\\ 0 & 1 + s & t_y\\ 0 & 0 & 1\end{bmatrix}\mathbf{x}_t
 \label{eq:PS}
\end{equation}

We utilize the Inverse Comompositional Spatial Transformer Networks (IC-STN) \cite{ICSTN} for stacking multiple warping blocks for better performance. We extend the work in \cite{ICSTN} to support pseduo-similarity and different warp types. A detailed study of which warp type performs the best is given in  Sec. \ref{subsec:AlgoDesign}. We further divide the experiments into table-top experiments  (Sec. \ref{sec:DeskExpts}) and flight experiments (Sec. \ref{sec:FlightExpts}).

\section{Table-top Experiments and Evaluation}
\label{sec:DeskExpts}


\subsection{Data Setup, Training and Testing Details}
\label{subsec:TrainTestDetails}
We train and test all our networks on the MS-COCO dataset \cite{MSCOCO}, using the \texttt{train2014} and \texttt{test2014} splits for training and testing. During training, we obtain a random crop of size 300 $\times$ 300 px. (denoted as $\mathcal{I}_t$ or $\mathcal{I}_1$), which is then warped using pseudo-similarity (2D translation + scale) to synthetically generate $\mathcal{I}_{t+1}$ or $\mathcal{I}_2$ obtained by using a random warp parameter in the range $\gamma_1 = \pm \begin{bmatrix} 0.25 & 0.20 & 0.20 \end{bmatrix}$ (unless specified otherwise). Then the center 128 $\times$ 128 px. patch are extracted (to avoid boundary effects) which are denoted as $\mathcal{P}_1$ and $\mathcal{P}_2$, respectively. A stack of $\mathcal{P}_1$ and $\mathcal{P}_2$ patches of size 128 $\times$ 128 $\times$ $2N_c$ ($N_c$ is the number of channels in $\mathcal{P}_1$ and $\mathcal{P}_2$, i.e., 3 if RGB 1 if grayscale) is fed into the network to obtain the predicted warp parameters: $\mathbf{\widetilde{h}} = \begin{bmatrix}s & t_x & t_y\end{bmatrix}^T$.

Our networks were trained in Python$^\text{TM}$ 2.7 using TensorFlow\footnote{\url{https://www.tensorflow.org/}} 1.14 on a Desktop computer (described in Sec. \ref{subsec:HWPlatforms}) running Ubuntu 18.04. We used a mini-batch-size of 32 for all the networks with a learning rate of 10$^{-4}$ without any decay. We trained all our networks using the ADAM optimizer for 100 epochs with early termination if we detect over-fitting on our validation set.

We tested our trained networks using two different configurations: (i) In-domain and (ii) Out-of-domain. In the in-domain testing (not to be confused with in-training dataset), we warped images from the \texttt{test2014} split of the MS-COCO dataset using a random warp parameter in range $\gamma_1 = \pm \begin{bmatrix} 0.25 & 0.20 & 0.20 \end{bmatrix}$ (same warp range as training). This was used for evaluation as described in Sec. \ref{subsec:EvalMetrics}. To commend on the generalization of the approach, we also tested it on out-of-domain warps, i.e., twice the warp range it was originally trained on, denoted by $\gamma_2 = \pm \begin{bmatrix} 0.50 & 0.40 & 0.40 \end{bmatrix}$. This test was  constructed to highlight the generalizability vs. speciality of the network configuration. Next we discuss the loss network architectures.

\subsection{Network Architectures}
We use five network architectures inspired by previous works. We call the networks as follows: VanillaNet \cite{VGGNet}, ResNet \cite{ResNet}, ShuffleNet \cite{ShuffleNet}, MobileNet \cite{MobileNet} and SqueezeNet \cite{SqueezeNet}. Each network is composed of various blocks. The output of each block is used as an incremental warp as proposed in \cite{ICSTN}. We use for each of these blocks  the same name as the network, for e.g., VanillaNet has VanillaNet blocks. We  use the following shorthand to denote the architecture. VanillaNet$_a$ denotes VanillaNet with $a$ VanillaNet blocks. We modify the networks from their original papers in the following manner: We exclude max-pooling blocks and replace them with stride in the previous convolutional block. Instead of varying sub-sampling rates (rates at which strides change with respect to depth of the network), we keep it fixed to the same value at every layer. All the architectures used are illustrated in Fig. \ref{fig:NetworkArchitectures}. The architectures shown in Fig. \ref{fig:NetworkArchitectures} are repeated for every warping block used, where each  block predicts the incremental warp as proposed in \cite{ICSTN}. We exclude the number of filters for the purpose of clarity, but they can be found in the code provided with the supplementary material which will be released upon the acceptance for publication.

We also test two different sizes of networks, i.e., large (model size $\le$ 8.3 MB) and small (Model size $\le$ 0.83 MB). Here the model sizes are computed for storing \texttt{float32} values for each neuron weight. Due to file format packing there is generally a small overhead and the actual file sizes are slightly larger.




\begin{figure*}[t!]
     \centering
     \includegraphics[width=\textwidth]{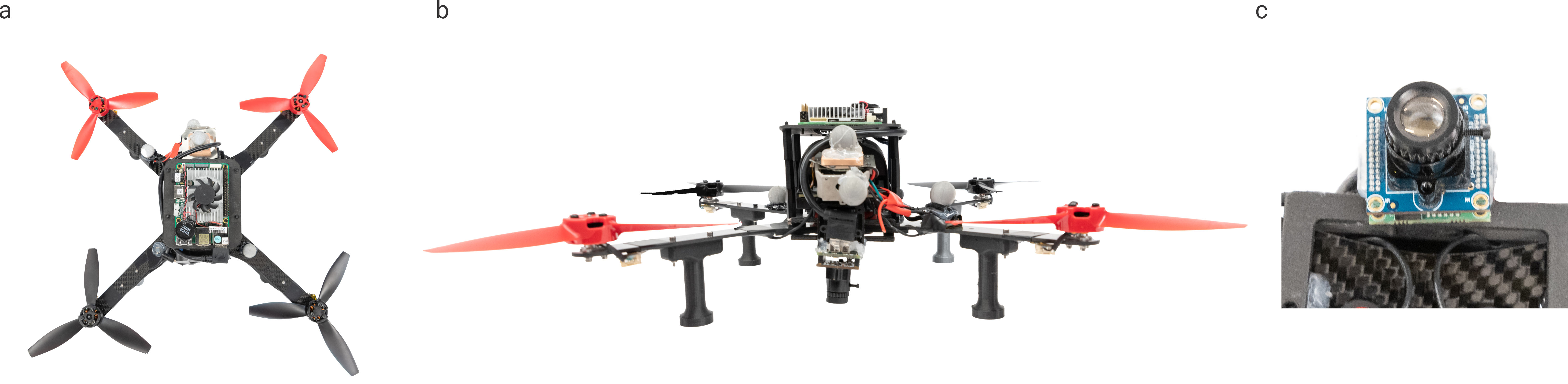}
     \caption{PRG Husky-$360\gamma$ platform used in flight experiments. (a) Top view, (b) front view, (c) down-facing leopard imaging camera.}
     \label{fig:PRGHusky}
 \end{figure*}

\subsection{Loss Functions}
The trivial way  to learn the warp parameters directly is to use  a supervised loss, since the labels are ``free'' as the data is synthetically generated on the fly. The loss function used in the supervised case is given as

\begin{equation}
    \mathcal{L}_s = \argmin_{\mathbf{\widetilde{h}}}\,\, \mathbb{E} \left(\Vert \mathbf{\widetilde{h}} - \mathbf{\hat{h}}\Vert_2\right),
    \label{eq:L2}
\end{equation}

where $\mathbf{\widetilde{h}},\,\, \mathbf{\hat{h}},\,\, \mathbb{E}$ are the predicted parameters, ideal parameters and expectation/averaging operator respectively. 
We also study the question ``Does using self-supervised loss give better out-of-domain generalization performance?'' The unsupervised losses are generally complicated since they are under-constrained compared to the one in supervised approaches. Hence, we study different unsupervised loss functions. We can write the loss functions as:

\begin{equation}
\mathcal{L}_{us} = \argmin_{\mathbf{\widetilde{h}}}\,\, \mathbb{E}\left(\mathcal{D}\left(\mathcal{W} \left(\mathcal{P}_{1},  \mathbf{\widetilde{h}}\right), \mathcal{P}_{2}\right)  + \lambda_i\mathcal{R}_i\right),
\end{equation}

where $\mathcal{W}$ is a generic differentiable warp function, which can take on different mathematical formulations based on it's second argument (model parameters), and $\mathcal{D}$ represents a distance measuring image similarity between the image frames. Finally, $\lambda_i,\,\, \mathcal{R}_i$ represent the $i^{\text{th}}$ Lagrange multiplier and it's corresponding regularization function. We experiment with different $\mathcal{D}$ and $\mathcal{R}$ functions described below.

\begin{align}
    \mathcal{D}_{\text{L1}}\left(A, B\right) &= \mathbb{E}\left(\Vert A - B\Vert_1\right)\\
     \mathcal{D}_{\text{Chab}}\left(A, B\right) &= \left(\left(\left(A - B\right)^2 + \epsilon^2\right)^\alpha\right)\\
    \mathcal{D}_{\text{SSIM}}\left(A, B\right) &= \mathbb{E}\left( \dfrac{1-\text{SSIM}\left( A, B\right)}{2}  + \alpha \left(\Vert A - B\Vert_1\right)\right)\\
    \mathcal{D}_{\text{Robust}}\left(A, B\right) &= \mathbb{E}\left(\dfrac{b}{d}\left(\left(\dfrac{\left(\sfrac{\left(A - B\right)}{c}\right)^{2}}{b} + 1 \right)^{\sfrac{d}{2}} - 1\right)\right)
\end{align}

\[  b = \Vert 2 - \hat{\alpha} \Vert_1 + \epsilon; \qquad
    d = \begin{cases}
    \hat{\alpha} + \epsilon\quad \text{if } \hat{\alpha} \ge 0\\
    \hat{\alpha} - \epsilon\quad \text{if } \hat{\alpha} < 0\\
    \end{cases}
    \]
\[ \hat{\alpha}_i = \left(2 - 2\epsilon_{\alpha}\right) \dfrac{e^{\alpha_i}}{e^{\alpha_i} + 1} \quad \forall i\]

Here, $\mathcal{D}_{\text{L1}}$ is the generic $l_1$ photometric loss \cite{L1Loss} commonly used for traditional images, $\mathcal{D}_{\text{Chab}}$ is the Chabonnier loss \cite{chabonnierloss} commonly used for optical flow estimation, $\mathcal{D}_{\text{SSIM}}$ is the loss based on Structure Similarity  \cite{SSIM} commonly used for learning ego-motion and depth from traditional image sequences and $\mathcal{D}_{\text{Robust}}$ is the robust loss function presented in \cite{robustloss}. Also, note that $\alpha$ has different connotation in each loss function. 


The functions given above can take any generic input such as the raw image or a function of the image. In our paper we experiment with different inputs such as the raw RGB image, grayscale image, high-pass filtered image and image cornerness score (denoted by $\mathcal{I}$, $\mathcal{G}$, $\mathcal{Z(I)}$ and $\mathcal{C(I)}$ respectively). The same set of functions can be used both as the metric function and the regularization function. We will denote the combination using a shorthand representation. Consider using the loss function with SSIM on raw images as the metric function and photometric L1 on high-pass filtered images as the regularization function with a Lagrange multiplier of $5.0$, the shorthand for this function is given in Eq. \ref{eq:LossShorthand}.


\begin{equation}
   \mathcal{D}_{\text{SSIM}}\left(\mathcal{I}\right) + 5.0 \mathcal{D}_{\text{L1}}\left(\mathcal{Z(I)}\right)
   \label{eq:LossShorthand}
\end{equation}

\subsection{Evaluation Metrics}
\label{subsec:EvalMetrics}
We use the following evaluation metrics to quantify the performance of each network. Let the predicted warp parameters be $\mathbf{\widetilde{h}} = \begin{bmatrix}\widetilde{s} & \widetilde{t_x} & \widetilde{t_y}\end{bmatrix}^T$ and the ideal warp parameters be $\mathbf{\hat{h}} = \begin{bmatrix}\hat{s} & \hat{t_x} & \hat{t_y}\end{bmatrix}^T$. $W$ and $H$ denote the image width and height respectively. Then the scale and translation error in pixel are given as:

\begin{align}
    \mathcal{E}_{\text{scale}} &= \mathbb{M}\left( \sqrt{\dfrac{W^2 + H^2}{2}}\vert \widetilde{s} - \hat{s} \vert \right)\\
    \mathcal{E}_{\text{trans}} &= \mathbb{M}\left( \dfrac{\sqrt{\left(W\left( \widetilde{t_x} - \hat{t_x} \right) \right)^2 + \left(H\left( \widetilde{t_y} - \hat{t_y} \right) \right)^2}}{2} \right)
\end{align}

Here, $\mathbb{M}$ denotes the median value (we choose the median value over the mean to reject  outlier samples with low texture).

We also convert errors to accuracy percentage as follows:

\begin{equation}
    \mathcal{A} = \left( 1 - \dfrac{\mathcal{E}_{\text{scale}} + \mathcal{E}_{\text{trans}}}{\mathbb{I}_{\text{scale}} + \mathbb{I}_{\text{scale}}}\right) \times 100\%
    \label{eq:Acc}
\end{equation}

Here, $\mathbb{I}_{\text{scale}}$ and $\mathbb{I}_{\text{trans}}$ denote the identity errors for scale and translation respectively (error when the prediction values are zero).


\begin{table*}[t!]
\centering
\caption{Different Computers Used on Aerial Robots.}
\resizebox{0.95\textwidth}{!}{
\label{tab:HardwarePlatforms}
\begin{tabular}{llllllllll}
\toprule
\multirow{2}{*}{Name} & Cost \multirow{2}{*}{$\quad\downarrow$} & Size \multirow{2}{*}{$\qquad\qquad\quad\downarrow$} & Weight \multirow{2}{*}{$\downarrow$}  & CPU & CPU Clock Speed \multirow{2}{*}{$\uparrow$} & RAM \multirow{2}{*}{$\uparrow$} & GPU \multirow{2}{*}{$\quad\uparrow$} & Max. Power$^*$ \multirow{2}{*}{$\downarrow$} & \multirow{2}{*}{Ease of use} \multirow{2}{*}{$\uparrow$}\\
& (USD) & ($l\times w \times h$ mm) & (g) &  Arch. & $\times$ Threads (GHz) & (GB) &  (GOPs) &  (W)\\
\hline
NanoPi Neo Core 2 LTS\footnotemark & 28 & $\mathbf{40 \times 40 \times 3}$ & \textbf{7} & ARM & 1.36 $\times$ 4 & 1.0 & - & \textbf{10} & \faStar \faStar \faStar\\
BananaPi M2-Zero\footnotemark & \textbf{24} & $65 \times 30 \times 5$ & 15 & ARM & 1.00 $\times$ 4 & 1.0 &  - & \textbf{10} & \faStar \faStar \faStar\\
Google Coral Dev Board\footnotemark & 150 & $88 \times 60 \times 22$ & 136 & ARM & 1.50 $\times$ 4 & 1.0 & 4000 & \textbf{10} & \faStar \faStar \faStar\faStarHalf \\
Google Coral Accelerator\footnotemark & 75 & $65 \times 30 \times 8$ & 20 & - & - & - & 4000 & 4.5 & \faStar \faStar \faStar \faStar\\
NVIDIA$^\text{\textregistered}$ Jetson$^\text{TM}$ TX2\footnotemark & 600 & $87 \times 50 \times 48$ & 200 & ARM & 2.00 $\times$ 6 & 8.0 & 1500 & 15 & \faStar \faStar \faStar \\
Intel$^\text{\textregistered}$ Up Board\footnotemark  & 159 & $86 \times 56 \times 20$ & 98 & x86 & 1.92 $\times$ 4 & 1.0 &  - & \textbf{10} & \faStar \faStar \faStar \faStar \faStarHalf\\
Laptop\footnotemark & 1600 & $391 \times 267 \times 31$ & 2200 & x86 & 3.40 $\times$ 8 & 16.0 & 6463 & 180 & \faStar \faStar \faStar \faStar \faStar \\
\bottomrule
$^*$Power consumption is for board not for chip.
\end{tabular}}
\end{table*}
\footnotetext[2]{\url{http://nanopi.io/nanopi-neo-core2.html}}
\footnotetext[3]{\url{http://www.banana-pi.org/m2z.html}}
\footnotetext[4]{\url{https://coral.ai/products/dev-board/}}
\footnotetext[5]{\url{https://coral.ai/products/accelerator}}
\footnotetext[6]{\url{https://www.nvidia.com/en-us/autonomous-machines/embedded-systems/jetson-tx2/}}
\footnotetext[7]{\url{https://up-board.org/}}
\footnotetext[8]{\url{https://www.asus.com/us/ROG-Republic-Of-Gamers/ROG-GL502VS/}}
\footnotetext[9]{\url{https://www.ibuypower.com/}}

\subsection{Hardware Platforms}
\label{subsec:HWPlatforms}
We tested multiple small factor computers and Single Board Computers (SBCs) (we'll refer to both as computers or computing devices or hardware) which are commonly used on aerial platforms. The platforms tested are summarized in Table \ref{tab:HardwarePlatforms}. Details not present in the table are explained next.

We overclocked the NanoPi Neo Core 2 LTS to 1.368 GHz from the base clock of 1.08 GHz to obtain better performance. This necessitated the use of an active cooling solution for long duration operation (more than 3 mins continuously). The weight of the cooling solution (4 g) and micro-SD card (0.5 g) are not included in Table \ref{tab:HardwarePlatforms}. Without the active cooling solution the computer goes into a thermal shutdown which will be harmful during in-flight operation. One can change the clock speed to certain available frequencies between 0.48 GHz and 1.368 GHz. To run larger models on the NanoPi's measly 1 GB of RAM we allocated 1 GB of SWAP memory on the Sandisk UHC-II class-10 microSD card which has much lower transfer speed as compared to RAM. The same cooling and SWAP solution were used for BananaPi M2-Zero as well. Note that, the Google Coral USB Accelerator is not officially supported with the NanoPi but we discovered a workaround which will be released in the accompanying supplementary material. Also, the Google Coral USB Accelerator is attached to a USB 2.0 port which limits its maximum performance when used with the NanoPi. The only reason why one would use BananaPi over the NanoPi is for the smaller width (30 mm versus 40 mm) which could be suited for a non X shaped quadrotor. However, the NanoPi is lighter, faster and has less area as compared to the BananaPi. Both NanoPi and BananaPi ran Ubuntu 16.04 LTS core with TensorFlow 1.14.

A significant speedup of upto 576$\times$ were obtained on the Coral Dev and the Coral USB accelerator when the original TensorFlow model was converted into TensorFlow-Lite and optimized for Edge TPU compilation. Without the Edge-TPU optimization the models run on the CPUs of these computers are far slower than the Tensor cores. To use both Coral Dev and Coral USB accelerator TensorFlow-Lite-Runtime 2.1 is used. The Coral Dev board ran Mendel Linux 1.5.

The NVIDIA$^\text{\textregistered}$ Jetson$^\text{TM}$ TX2 in our setup is used with the Connect Tech's Orbitty carrier board (weighing 41 g and its weight is included in Table \ref{tab:HardwarePlatforms}). This carrier board allows for the most compact setup with the TX2. Note that the NVIDIA$^\text{\textregistered}$ Jetson$^\text{TM}$ TX2 can be used without the active heatsink for a short duration (less than 5 mins) reducing its weight by 59 g (a massive 30\% reduction in weight without any loss of performance). However, extended operation without the active heatsink can result in thermal shutdown or permanent damage to the computer. During our experiments, we set the operation mode to fix the CPU and GPU frequencies to the maximum available value, and this in-turn maxes out the power consumption (the steps to achieve this will be released in the accompanying supplementary material). The TX2 ran  Linux For Tegra (L4T) R28.2.1 installed using Jetpack 3.4 with TensorFlow 1.11.

We use neither the AVX (not supported on hardware) nor the SSE (not supported by the TensorFlow version supported on Up board) instruction set on the Intel$^\text{\textregistered}$ Up board. We speculate huge speed-ups if a future version of TensorFlow supports SSE on the Up board. The Up board ran Ubuntu 16.04 LTS with TensorFlow 1.11.

The laptop is an Asus ROG GL502VS with Intel$^\text{\textregistered}$ Core$^\text{TM}$ i7-6700HQ and GTX 1070 GPU and weighs about 2200 g which can be used on a large ($\ge$ 650 mm sized) quadrotor. Also, note that an Intel NUC coupled to an eGPU can also be used with similar specifications but this setup will still be heavier, arguably more expensive and probably less reliable than gutting out a gaming laptop. The laptop ran Ubuntu 18.04 LTS with TensorFlow 1.14.

The desktop PC is a custom built full tower PC from iBUYPOWER with an Intel$^\text{\textregistered}$ Core$^\text{TM}$ i9-9900KF and NVIDIA$^\text{\textregistered}$ Titan-Xp GPU which cannot be used on an aerial robot but is included to  serve as a reference. The desktop PC ran Ubuntu 18.04 LTS with TensorFlow 1.14. Whenever possible we use the NEON SIMD instruction set for ARM computers and the SSE instructions for x86 systems.


\section{Flight Experiments and Evaluation}
\label{sec:FlightExpts}
This section presents real-world experiments flying various simple trajectories with odometry estimation using PRGFlow. We use the lowest avg. pixel error 8.3 MB (large) model for evaluation (ResNet$_4$ with T$\times$2, S$\times$2 warping configuration). The network outputs a  $\mathbb{R}^{3 \times 1}$ vector denoted as $\mathbf{\widetilde{h}} = \begin{bmatrix}\widetilde{s} & \widetilde{t_x} & \widetilde{t_y}\end{bmatrix}^T$. The predicted translational pixel velocities (global optical flow) $\begin{bmatrix} \widetilde{t_x} & \widetilde{t_y}\end{bmatrix}^T$ are converted to real world velocities by scaling them using the focal length and the depth (adjusted altitude) \cite{px4flow}. A similar treatment is given to the $Z$-pixel velocity $s$. Also, we obtain attitude using a Madgwick filter \cite{Madgwick} from a 9-DoF IMU, which is used to remove rotation between consecutive frames, and feed the values  into the network. Finally, to obtain odometry, we simply perform dead-reckoning on the velocities obtained by our network. \textit{Our networks were trained on MS-COCO as described in \ref{subsec:TrainTestDetails}, and are used  in the flight experiments without any fine-tuning or re-training to highlight the generalizability of our approach}. Note that the performance can be significantly boosted by multi-frame fusion of predictions, and we leave this  for future work.

\subsection{Experimental Setup}

We tested our algorithm on the PRG Husky-$360\gamma$ platform\footnote{\url{https://github.com/prgumd/PRGFlyt/wiki/PRGHusky}}: a modified version of the Parrot$^\text{\textregistered}$ Bebop 2 for its ease of use which was originally designed for pedagogical reasons. It is equipped with a down facing Leopard Imaging LI-USB30-M021 global shutter camera\footnote{\url{https://leopardimaging.com/product/usb30-cameras/usb30-camera-modules/li-usb30-m021/}} with a 16 mm lens which gives a diagonal field of view of $\sim22^\circ$ (Refer to Fig. \ref{fig:PRGHusky}). The PRG Husky also contains an 9-DoF IMU and a down-facing sonar for attitude and altitude measurements respectively. Higher level control commands are given by an on-board companion computer: Intel$^\text{\textregistered}$ Up Board at 20 Hz. The images are recorded at $90$ Hz at a resolution of $640 \times 480$ px. The overall takeoff weight of the flight setup is $730$ g (which gives a thrust-to-weight ratio of $\sim1.5$) with diagonal motor to motor dimension of $360$ mm.

The flight experiments were conducted in the Autonomy Robotics and Cognition (ARC) lab's netted indoor flying space at the University of Maryland, College Park. The total flying volume is about $6 \times 5.5 \times 3.5$ m$^3$. A Vicon motion capture system with 12 vantage V8 cameras were used to obtain ground truth at 100 Hz.

The PRG Husky was tested on five trajectories: \texttt{circle, moon, line, figure8} and \texttt{square} which involve change in both attitude and altitude with an average velocity of about 0.5 ms$^{-1}$ and a maximum velocity of 1.5 ms$^{-1}$.

\begin{table}[t!]
\centering
\caption{Quantitative evaluation of different warping combination for Pseudo-similarity estimation.}
\resizebox{\columnwidth}{!}{
\label{tab:DiffWarping}
\begin{tabular}{lllllll}
\toprule
\multirow{2}{*}{Network (Warping)} & \multicolumn{2}{l}{$\mathcal{E}_{\text{scale}}$ (px.)} \multirow{2}{*}{$\downarrow$} &  \multicolumn{2}{l}{$\mathcal{E}_{\text{trans}}$ (px.)} \multirow{2}{*}{$\downarrow$}& \multirow{2}{*}{FLOPs (G) $\downarrow$} & Num. \multirow{2}{*}{$\qquad\,\,\downarrow$} \\
 & $\gamma_1$ & $\gamma_2$ & $\gamma_1$ & $\gamma_2$ & & Params (M) \\
\hline
Identity & 11.4 & 22.8 & 10.3 & 20.4 & --  & -- \\
VanillaNet$_1$ (PS$\times$1) & 2.4 & 15.0 & \textbf{1.3} & 12.5 & \textbf{0.37} & \textbf{2.07} \\
VanillaNet$_1$ (PS$\times$1) DA & 4.1 & 17.7 & 2.3 & 14.2  & \textbf{0.37} & \textbf{2.07} \\
VanillaNet$_2$ (PS$\times$2) & 2.2 & 9.9 & 1.4 & 12.4 & 0.42 & 2.17\\
VanillaNet$_2$ (S$\times$1, T$\times$1) & 2.5 & 15.2 & 1.5 & 12.2 & 0.46 & 2.10 \\
VanillaNet$_2$ (T$\times$1, S$\times$1) & 2.5 & 15.1 & 1.5 & 12.5 & 0.42 & 2.15\\
VanillaNet$_4$ (PS$\times$4) & 2.3 & 11.9 & 1.5 & 14.9 & 0.42 & 2.15\\
VanillaNet$_4$ (S$\times$2, T$\times$2) & 2.6 & 15.4 & 1.6 & 12.6 & 0.46 & 2.08\\
VanillaNet$_4$ (T$\times$2, S$\times$2) & \textbf{2.0} & 8.5 & 1.5 & 12.5 & 0.46 & 2.08 \\
VanillaNet$_4$ (T$\times$2, S$\times$2) $\gamma_2$ & 2.7 & \textbf{2.8} & 4.6 & \textbf{7.2} & 0.46 & 2.08\\
 \bottomrule
\end{tabular}}
\end{table}

\begin{table}[t!]
\centering
\caption{Quantitative evaluation of different network architectures for Pseudo-similarity estimation using T$\times$2, S$\times$2 warping block for large model ($\le$8.3 MB).}
\resizebox{0.8\columnwidth}{!}{
\label{tab:DiffArchLarge}
\begin{tabular}{lllllll}
\toprule
\multirow{2}{*}{Network} & \multicolumn{2}{l}{$\mathcal{E}_{\text{scale}}$ (px.)} \multirow{2}{*}{$\downarrow$} &  \multicolumn{2}{l}{$\mathcal{E}_{\text{trans}}$ (px.)} \multirow{2}{*}{$\downarrow$} & \multirow{2}{*}{FLOPs (G) $\downarrow$} & Num. \multirow{2}{*}{$\qquad\,\,\downarrow$} \\
 & $\gamma_1$ & $\gamma_2$ & $\gamma_1$ & $\gamma_2$ & & Params (M) \\
\hline
Identity & 11.4 & 22.8 & 10.3 & 20.4 & -- & --  \\
VanillaNet$_4$ & 1.9 & 6.4 & 1.5 & 12.4 & 0.46 & 2.08 \\
ResNet$_4$ & \textbf{1.7} & 15.1 & \textbf{0.9} & \textbf{10.1} & 0.59 & 2.12\\
SqueezeNet$_4$ & 2.1  & \textbf{5.7} & 2.2 & 13.8 & 2.20 & 2.12\\
MobileNet$_4$ & 4.0 & 14.2  &  1.6 & 12.0 & \textbf{0.41} & \textbf{2.04}  \\
ShuffleNet$_4$ & 6.4  & 17.4  & 3.0  & 13.9  & 1.20 & 2.10 \\
\bottomrule
\end{tabular}}
\end{table}

\begin{table}[t!]
\centering
\caption{Quantitative evaluation of different network architectures for Pseudo-similarity estimation using T$\times$2, S$\times$2 warping block for small model ($\le$0.83 MB).}
\resizebox{0.8\columnwidth}{!}{
\label{tab:DiffArchSmall}
\begin{tabular}{lllllll}
\toprule
\multirow{2}{*}{Network} & \multicolumn{2}{l}{$\mathcal{E}_{\text{scale}}$ (px.)} \multirow{2}{*}{$\downarrow$} &  \multicolumn{2}{l}{$\mathcal{E}_{\text{trans}}$ (px.)} \multirow{2}{*}{$\downarrow$}& \multirow{2}{*}{FLOPs (G) $\downarrow$} & Num. \multirow{2}{*}{$\qquad\,\,\downarrow$} \\
 & $\gamma_1$ & $\gamma_2$ & $\gamma_1$ & $\gamma_2$ & & Params (M)\\
\hline
Identity & 11.4 & 22.8 & 10.3 & 20.4 & -- & --  \\
VanillaNet$_4$ & 3.3 & 8.9 & 3.1 & 14.0 & 0.18 & 0.21 \\
ResNet$_4$ & 4.4 & 12.5 & \textbf{2.4} & \textbf{12.1} & 0.20 & \textbf{0.20}\\
SqueezeNet$_4$ & \textbf{2.4} & \textbf{5.6} & 4.0 & 14.9 & 0.19 & \textbf{0.20}\\
MobileNet$_4$ & 8.3 & 18.7 & 3.7 & 13.4 & 0.16 & \textbf{0.20} \\
ShuffleNet$_4$ & 8.3 & 17.6 & 4.6 & 15.7 & \textbf{0.13} & 0.21 \\
\bottomrule
\end{tabular}}
\end{table}

\begin{table}[h!]
\centering
\caption{Quantitative evaluation of different network inputs for Pseudo-similarity estimation using T$\times$2, S$\times$2 warping block for large model ($\le$8.3 MB).}
\resizebox{0.7\columnwidth}{!}{
\label{tab:DiffInputs}
\begin{tabular}{lllll}
\toprule
\multirow{2}{*}{Testing Data (Training Data)} & \multicolumn{2}{l}{$\mathcal{E}_{\text{scale}}$ (px.)} \multirow{2}{*}{$\downarrow$} &  \multicolumn{2}{l}{$\mathcal{E}_{\text{trans}}$ (px.)} \multirow{2}{*}{$\downarrow$} \\
 & $\gamma_1$ & $\gamma_2$ & $\gamma_1$ & $\gamma_2$  \\
\hline
Identity & 11.4 & 22.8 & 10.3 & 20.4 \\
$\mathcal{I}$ ($\mathcal{I}$) & 2.0 & 8.5 & \textbf{1.5} & 12.5 \\
$\mathcal{G}$ ($\mathcal{I}$)  & \textbf{1.8} & \textbf{6.3} & \textbf{1.5} & \textbf{12.3}\\
$\mathcal{G}$ ($\mathcal{G}$) & 2.7 & 14.1 & 1.6 & 12.7 \\
$\mathcal{Z(I)}$ ($\mathcal{Z(I)}$) & 13.1 & 9.4 & 10.4 & 16.0 \\
$\mathcal{G}$ ($\mathcal{Z(I)}$) & 11.8  & 20.7 & 9.8 & 19.8\\
$\mathcal{I}$ ($\mathcal{Z(I)}$) & 13.1 & 22.5 & 10.5 & 20.1 \\
$\mathcal{Z(I)}$ ($\mathcal{G}$) & 8.5 & 19.8 & 4.1 & 17.6\\
$\mathcal{Z(I)}$ ($\mathcal{I}$) & 17.2 & 20.1 & 4.2 & 17.4\\
\bottomrule
\end{tabular}}
\end{table}

\begin{table}[h!]
\centering
\caption{Quantitative evaluation of different loss functions for Pseudo-similarity estimation using PS$\times$1 warping block for large model ($\le$8.3 MB).}
\resizebox{\columnwidth}{!}{
\label{tab:DiffLossFunc}
\begin{tabular}{lllllll}
\toprule
\multirow{2}{*}{Loss Function (Architecture)} & \multicolumn{2}{l}{$\mathcal{E}_{\text{scale}}$ (px.)} \multirow{2}{*}{$\downarrow$} &  \multicolumn{2}{l}{$\mathcal{E}_{\text{trans}}$ (px.)} \multirow{2}{*}{$\downarrow$}& \multirow{2}{*}{FLOPs (G) $\downarrow$} & Num. \multirow{2}{*}{$\qquad\,\,\downarrow$} \\
 & $\gamma_1$ & $\gamma_2$ & $\gamma_1$ & $\gamma_2$ & & Params (M) \\
\hline
Identity & 11.4 & 22.8 & 10.3 & 20.4 & --  & -- \\
Supervised $\mathcal{L}_s$ (VanillaNet$_1$) & 2.4 & 15.0 & 1.3 & 12.5 & 0.37 & 2.07\\
$\mathcal{D}_{\text{Robust}}\left(\mathcal{I}, \mathcal{C(I)}\right)$ (ResSqueezeNet$_1$) & 12.9 & 25.2 & 7.2 & \textbf{11.7} & \textbf{1.01} & \textbf{2.18}\\
$\mathcal{D}_{\text{SSIM}}\left(\mathcal{I}\right)$ (ResSqueezeNet$_1$) & 3.4 & 21.2 & 6.0 & 13.8 & \textbf{1.01} & \textbf{2.18}\\
$\mathcal{D}_{\text{SSIM}}\left(\mathcal{I}\right) + 0.1\mathcal{D}_{\text{L1}}\left(\mathcal{C(I)}\right)$ (ResSqueezeNet$_1$)  & \textbf{2.0} & \textbf{16.1} & 6.2 & 14.6 & \textbf{1.01} & \textbf{2.18}\\
$\mathcal{D}_{\text{SSIM}}\left(\mathcal{I}\right) + 0.1\mathcal{D}_{\text{L1}}\left(\mathcal{Z(I)}\right)$ (ResSqueezeNet$_1$) & 2.7 & 16.6 & 6.4 & 13.6 & \textbf{1.01} & \textbf{2.18}\\
$\mathcal{D}_{\text{L1}}\left(\text{DB}\left(\mathpzc{E}\right)\right)$ \cite{evdodgenet}  & 5.4 & 17.7 & 3.7 & 16.5 & 4.92 & 3.6 \\
$\mathcal{D}_{\text{Chab}}\left(\text{DB}\left(\mathpzc{E}\right)\right)$ \cite{evdodgenet}  & 5.1 & 17.1 & 3.4 & 16.7 & 4.92 & 3.6 \\
Supervised $\text{DB}\left(\mathpzc{E}\right)$ \cite{evdodgenet}  & 4.1 & 16.2 & \textbf{3.3} & 15.1 & 4.92 & 3.6 \\
\bottomrule
\end{tabular}}
\end{table}

\begin{table}[h!]
\centering
\caption{Quantitative evaluation of different compression methods for Pseudo-similarity estimation using PS$\times$1 warping.}
\resizebox{\columnwidth}{!}{
\label{tab:DiffCompression}
\begin{tabular}{lllllllll}
\toprule
\multirow{2}{*}{Method} & \multicolumn{2}{l}{$\mathcal{E}_{\text{scale}}$ (px.)} \multirow{2}{*}{$\downarrow$} &  \multicolumn{2}{l}{$\mathcal{E}_{\text{trans}}$ (px.)} \multirow{2}{*}{$\downarrow$} & \multicolumn{2}{l}{$\mathcal{E}_{\text{scale}}$ (DA) (px.)} \multirow{2}{*}{$\downarrow$} &  \multicolumn{2}{l}{$\mathcal{E}_{\text{trans}}$ (DA) (px.)} \multirow{2}{*}{$\downarrow$}  \\
 & $\gamma_1$ & $\gamma_2$ & $\gamma_1$ & $\gamma_2$ & $\gamma_1$ & $\gamma_2$ & $\gamma_1$ & $\gamma_2$  \\
\hline
Identity & 11.4 & 22.8 & 10.3 & 20.4 & 11.4 & 22.8 & 10.3 & 20.4 \\
Teacher from Scratch  & 2.4 & 15.0 & 1.3 & 12.5 & 4.3 & 17.2 & 2.5 & 14.1 \\
Student from Scratch & \textbf{3.5} & \textbf{9.9} & 2.8 & 13.2 & \textbf{4.2} & \textbf{16.8} & 4.2 & 16.0\\
Projection Loss Student \cite{ProjectionNet} & 3.7 & 10.9 & 2.8 & \textbf{13.1} & 8.0 & 17.7 & 4.2 & \textbf{15.2}\\
Model Distillation Student \cite{TeacherStudent} & 3.8 & 12.3 & \textbf{2.7} & 13.4 & 7.2 & 17.7 & \textbf{4.0} & \textbf{15.2}\\
\bottomrule
\end{tabular}}
\end{table}

\begin{table}[h!]
\centering
\caption{Comparison of PRGFlow with different classical methods.}
\resizebox{\columnwidth}{!}{
\label{tab:TradMethods}
\begin{tabular}{llllll}
\toprule
Method & $\mathcal{E}_{\text{scale}}$ (px.) $\downarrow$ &  $\mathcal{E}_{\text{trans}} $ $\downarrow$ (px.) & $\mathcal{E}_{\text{scale}}$ (DA) (px.) $\downarrow$ &  $\mathcal{E}_{\text{trans}} $ (DA) (px.) $\downarrow$ & Time (ms) $\downarrow$\\
\hline
Identity & 11.4 & 10.3 & 11.4 & 10.3 & --  \\
Supervised $\mathcal{L}_s$ & 1.9 & 1.5 & \textbf{4.1} & 2.3 & \textbf{1.1}$^*$ \\
FFT Aligment \cite{reddy1996fft} & \textbf{0.3} & \textbf{0.1} & 13.4 & 6.0 & 35.5\\
SURF \cite{bay2006surf} & 0.4 & \textbf{0.1} & 11.2 & \textbf{1.3} & 17.6 \\
ORB \cite{rublee2011orb} & 0.6  & \textbf{0.1} & 11.5 & 1.4 &12.9 \\
FAST \cite{viswanathan2009features} & 0.8  & 0.2 & 12.0 & \textbf{1.3} & 55.9  \\
Brisk \cite{leutenegger2011brisk} & 0.7  & 0.2 & 13.0 & 1.4 & 38.7 \\
Harris \cite{harris1988combined} & 0.4 & \textbf{0.1} &  10.9 & 1.4 & 60.2  \\
\bottomrule
$^*$ On all cores.
\end{tabular}}
\end{table}

\section{Discussion}
\subsection{Algorithmic Design}
\label{subsec:AlgoDesign}
We answer the following questions in this section:
\begin{enumerate}
    \item What is the best warp combination?
    \item What network architecture is the best? Does the best network architecture vary with respect to the number of parameters?
    \item How does the input affect performance of a network?
    \item What is the best way to compress a model?
    \item When is it advisable to choose deep learning for odometry estimation?
\end{enumerate}

   \begin{figure}
     \centering
     \includegraphics[width=\columnwidth]{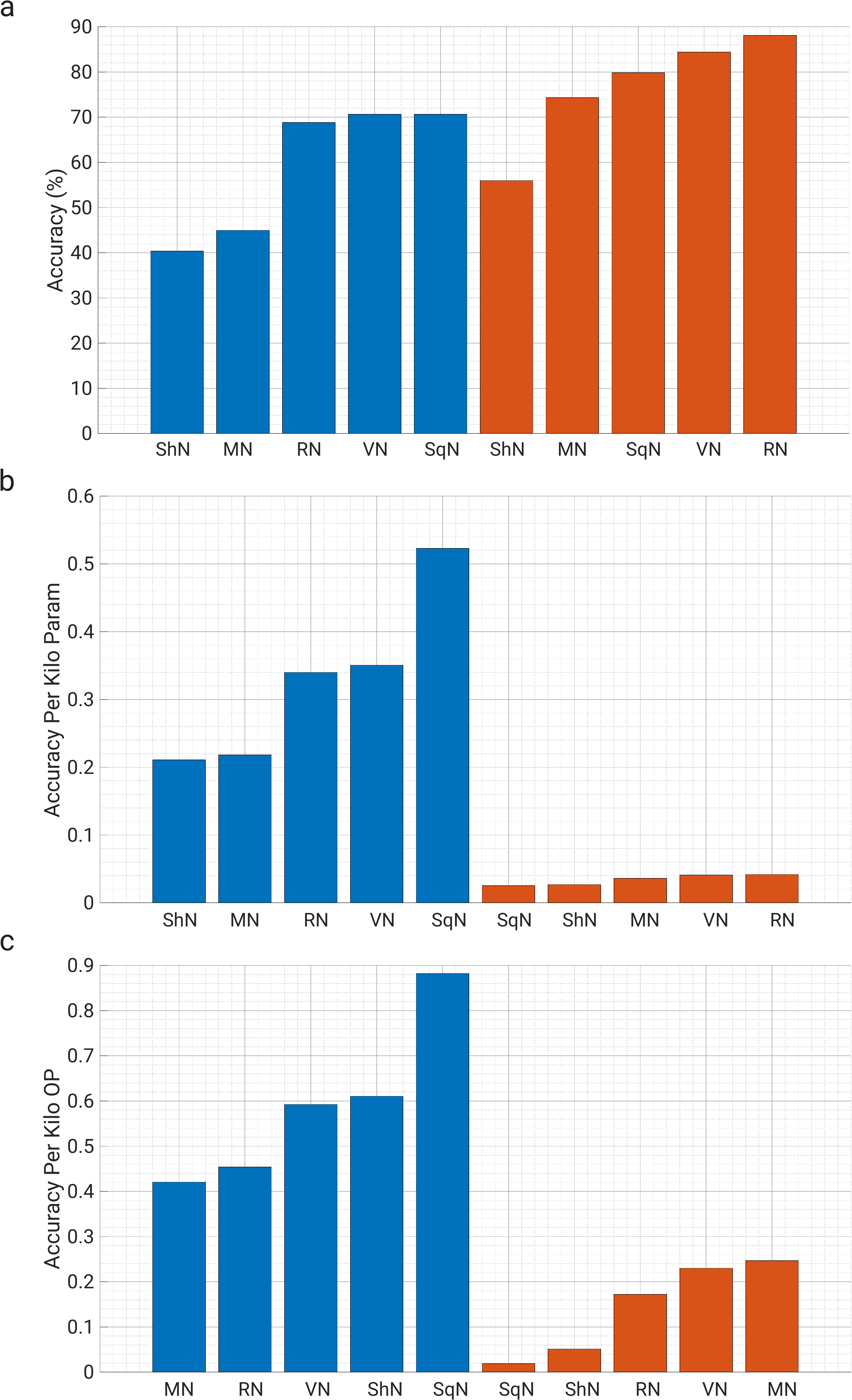}
     \caption{(a) Accuracy, (b) Accuracy per Kilo param, (c) Accuracy per Kilo OP for different network architectures. Blue and orange histograms denote small ($\le$0.83 MB) and large ($\le$8.3 MB) networks respectively. Here the following shorthand is used for network names: VN: VanillaNet, RN: ResNet, SqN: SqueezeNet, MN: MobileNet and ShN: ShuffleNet. All networks use T$\times$2, S$\times$2 warping configuration.}
     \label{fig:AccCombined}
 \end{figure}

The performance of different IC-STN warp combinations for VanillaNet are given in Table \ref{tab:DiffWarping}. The in-domain and out-of-domain test results have a headings of $\gamma_1$ and $\gamma_2$ respectively. All the networks were trained using supervised $l_2$ loss  (Eq. \ref{eq:L2}). Note that, the networks were constrained  to be $\le$8.3 MB. Under this condition, one can clearly observe the following trend when only using PS blocks for warping -- as the number of warping blocks increases the performance reaches a maximum and then deteriorates. This is because the number of neurons per warp block directly affects the performance, and increasing the number of warp blocks without increasing model size hurts accuracy when the number of neurons per warp block become small. An interesting observation is that predicting translation before scale almost always results in better performance and also decoupling the predictions of scale and translation using S and T blocks generally results in better performance (lower pixel error) as compared to predicting them together using PS blocks. This is serendipitous. As drift in the X and Y position  are generally higher, one could  obtain  these positions faster (since only a part of the network is used for output) and at higher frequency than the Z position \cite{thurstmixing}.

We also observe that training and testing with data augmentation of brightness, contrast, hue, saturation and Gaussian noise worsens the average performance by 73\% (comparing rows 2 and 3 in Table \ref{tab:DiffWarping}) since the network now has to learn to be agnostic to a myriad of image changes. Also, training on $\gamma_2$ (last two rows of Table \ref{tab:DiffWarping}) decreases the average error by 62\% when testing on $\gamma_2$ - we speculate it will further decrease with  increasing model size. This shows that as the deviation range in training increases, one can require more parameters for the same average performance. \textit{Overall the best performing warp configuration is T$\times$2, S$\times$2 when considering an average of both in-domain and out-of-domain tests.}

Next, let us study the performance with different network architectures. We use  T$\times$2, S$\times$2 warps for all the networks, and they are trained using supervised $l_2$ loss  (Eq. \ref{eq:L2}). The results for networks constrained by model size $\le$8.3 MB and $\le$0.83 MB can be found in Tables \ref{tab:DiffArchLarge} and  \ref{tab:DiffArchSmall} respectively. For ease of analysis, the results are also visually depicted in Fig. \ref{fig:AccCombined}. One can clearly observe that ResNet gives the best performance for both small and large networks (Fig. \ref{fig:AccCombined}{\color{red}a}) and should be the network architecture of choice when designing for maximizing accuracy without any regard to the number of parameters or amount of OPs (operations). However, if one has to prioritize maximizing accuracy whilst minimizing number of parameters, then SqueezeNet and ResNet would be the choice for smaller and larger networks respectively (Fig. \ref{fig:AccCombined}{\color{red}b}). Another trend one can observe is that one would need 10$\times$ more parameters for a 19\% increase in accuracy. Lastly, if one has to prioritize in maximizing accuracy whilst minimizing the number of OPs, then SqueezeNet and MobileNet would be the choice for smaller and larger networks respectively (Fig. \ref{fig:AccCombined}{\color{red}c}). Clearly, the most optimal architecture in-terms of accuracy, number of parameters and OPs  is SqueezeNet for smaller networks and ResNet for  larger networks. \textit{The effect of the choice of network architecture is fairly significant on the accuracy, number of parameters and OPs and needs to be carefully considered when designing a network for deployment on an aerial robot}. Also, note that sometimes using a classical approach to solve a small part of the problem can significantly simplify the learning problem for the network thereby maximizing accuracy and minimizing the number of parameters and OPs \cite{gapflyt}.

To gather more insight into what data representation is more important for odometry estimation, we explore training and testing on the following data representations: (a) RGB image, (b) Grayscale image, (c) High pass filtered image. We choose the T$\times$2, S$\times$2 warp configuration and the VanillaNet$_4$ architecture trained using
supervised $l_2$ loss  constrained by model size $\le$8.3 MB. Table \ref{tab:DiffInputs} summarizes the results obtained. Surprisingly, training on RGB images and evaluating on RGB images gives worse performance in the  testing both for in-domain ($\gamma_1$) and out-of-domain ($\gamma_2$) ranges than testing on grayscale data. We speculate that this is due to conflicting information in multiple channels. Another surprising observation is that training and/or testing on high-pass filtered images results in large errors, which is contrary to the  classical approaches. We speculate that this is because conventional neural networks rely on ``staticity'' of the image pixels (image pixels change slowly and are generally smooth).

We also explore the state-of-the-art self-supervised/unsupervised loss functions to test for claims of better out-of-domain generalization. We choose half-sized ($\le$4.15 MB) ResNet (to output scale) and SqueezeNet (to output translation) denoted as ResSqueezeNet trained using PS$\times$1 blocks for this experiment since it empirically gave us the best results (we exclude other architecture results for the purpose of brevity). We also include results from a VanillaNet$_1$ trained using PS$\times$1 blocks using the supervised $l_2$ loss function to serve as a reference (Refer to Table \ref{tab:DiffLossFunc}). We observe that scale error when using SSIM and cornerness (obtained using a heatmap as the output of \cite{Superpoint}) in the loss approaches the performance of the supervised network, but the translation error is almost three times that of the supervised network. Surprisingly, the supervised network also performs better than most unsupervised networks on out-of-domain tests. \textit{This hints that we need better loss functions for unsupervised methods and better network architectures to take advantage of these unsupervised losses.} From a practitioner's point of view, the supervised networks perform better and generalize better to out-of-domain. Another keen insight is that focusing on crafting better supervised loss functions may lead to a massive boost in performance.

Finally, we explore different strategies to compress the network. We specifically consider a setup where we compress a 8.3 MB model VanillaNet$_1$ PS$\times$1  to a 0.83 MB model VanillaNet$_1$ PS$\times$1. We test three different methods, (a) direct dropping of weights, (b) projection inspired loss and (c) model distillation. In the first method, we reduce the number of neurons and number of blocks to reduce the number of weights and train the smaller network from scratch. In the second method, we train both the larger and smaller networks together as given by Eq. \ref{eq:projloss} \cite{ProjectionNet}. Here, $\widetilde{h_T}, \widetilde{h_S}$ denote the predictions from the teacher and student network respectively. We choose $\lambda_1, \lambda_2, \lambda_3 $ as 1.0, 1.0 and 0.1 respectively. In the third method, we use an already trained teacher network (large 8.3 MB model) and define the loss to learn the predictions of the teacher using the student (small 0.83 MB model) as given by Eq. \ref{eq:TSloss} \cite{TeacherStudent}. When no data augmentation is used, we observe that directly dropping weights gives the best performance (Refer to Table \ref{tab:DiffCompression}). However, when the data augmentation is added the model distillation gives the best results, albeit only slightly better than directly dropping of weights. This observation is contrary to that observed with classification networks where massive boosts in performance are observed when using either Eq. \ref{eq:projloss} or \ref{eq:TSloss}. From a practitioner's point of view, the simplest method of directly dropping weights work the best for regression networks like the one used in this work and can provide up to 10$\times$ savings in the  number of parameters and OPs at the cost of $\sim$19\% accuracy. The same effect is observed when training and testing with and without data augmentation.

\begin{equation}
    \mathcal{L}_{\text{Proj}} = \lambda_1 \mathcal{L}_s\left(\hat{h}, \widetilde{h_T}\right) +  \lambda_2 \mathcal{L}_s\left(\hat{h}, \widetilde{h_S}\right) +  \lambda_3 \mathcal{L}_s\left(\widetilde{h_T}, \widetilde{h_S}\right)
    \label{eq:projloss}
\end{equation}

\begin{equation}
    \mathcal{L}_{\text{TS}} = \mathcal{L}_s\left(\widetilde{h_T}, \widetilde{h_S}\right)
    \label{eq:TSloss}
\end{equation}

To address the elephant in the room, we try to answer the following question: ``When should one use deep learning over a traditional approach?'' We compare the proposed deep learning approach PRGFlow to the common method of fast feature matching on aerial robots in Table \ref{tab:TradMethods}. Note that the runtime for traditional methods are given for \textsc{Matlab}\footnote{\url{https://www.mathworks.com/products/matlab.html}} implementations on one thread of the Desktop PC to standardize the libraries and optimizations used. Up to 5$\times$ and 10$\times$ speedup can be obtained using efficient C++ implementations on a multi-threaded CPU and GPU respectively (we don't explore this in our work). One can clearly observe that even with  C++ implementations the traditional handcrafted features are slower than the deep learning methods which can utilize the parallel hardware accelerations on GPUs. Though on the surface it seems like the traditional methods give far superior performance in terms of accuracy (lower error), the efficacy of deep learning approaches are brought into limelight when we train and test with data augmentations.  This simulates a bad quality camera common on smaller aerial robots. The drop in accuracy (from no-noise data to noisy data) in deep learning approaches is much less than that compared to traditional approaches, i.e., \textit{deep learning approaches on an average fail more gracefully  compared to traditional approaches.}

  \begin{figure*}
     \centering
     \includegraphics[width=\textwidth]{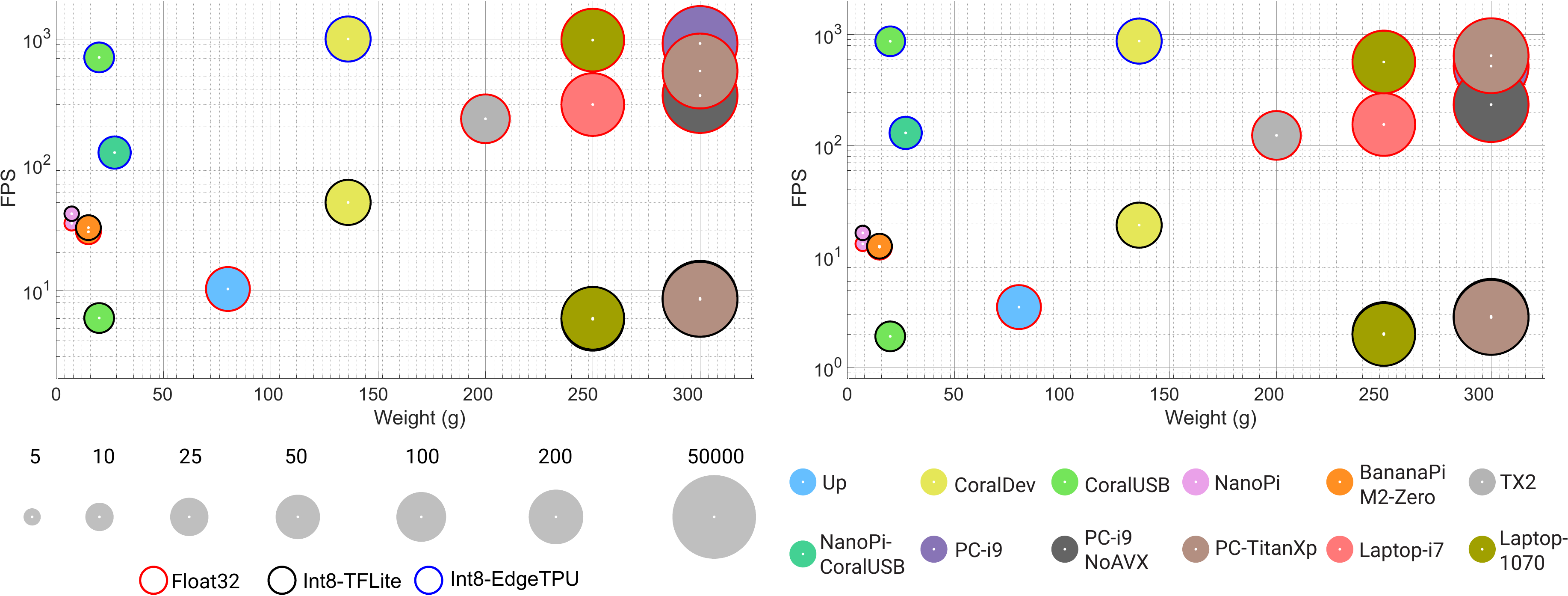}
     \caption{Weight vs. FPS for VanillaNet$_4$ (T$\times$2, S$\times$2) on different hardware and software optimization combinations. Left: small ($\le$0.83 MB) model, right: large ($\le$8.3 MB) model. The radius of each circle is proportional to log of volume of each hardware (this is shown in the legend below the plots with volume indicated on top of each legend in cm$^3$). The outline on each sample indicates the configuration of quantization or optimization used (\texttt{Float32} (red outline) is the original TensorFlow model without any quantization or optimization, \texttt{Int8-TFLite} (black outline) is the TensorFlow-Lite model with 8-bit Integer quantization and \texttt{Int8-EdgeTPU} (blue outline) is the TensorFlow-Lite model with 8-bit Integer quantization and Edge-TPU optimization. The samples are color coded to indicate the computer it was run on (shown in the legend on the bottom).  \textit{Also note that, Laptop and PC (Deskop) weight and volume values are not to actual scale for visual clarity in all images. All the figures in this paper use the same legend and color coding for ease of readability.}}
     \label{fig:VanillaNet}
 \end{figure*}

   \begin{figure*}
     \centering
     \includegraphics[width=\textwidth]{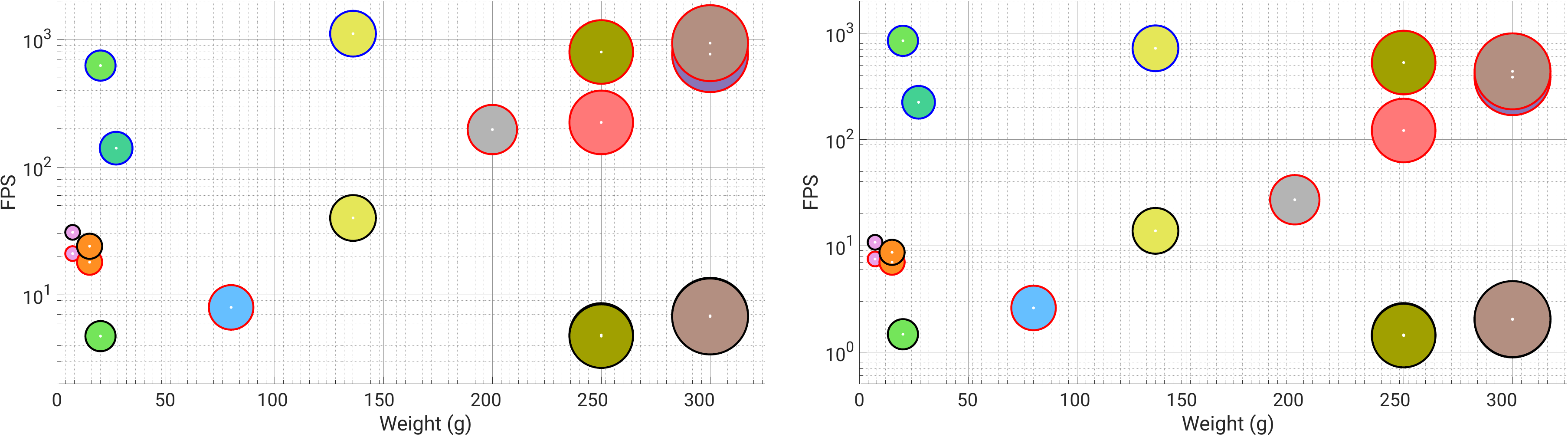}
     \caption{Weight vs. FPS for ResNet$_4$ (T$\times$2, S$\times$2) on different hardware and software optimization combinations. Left: small ($\le$0.83 MB) model, right: large ($\le$8.3 MB) model. The radius of each circle is proportional to log of volume of each hardware.}
     \label{fig:ResNet}
 \end{figure*}

   \begin{figure*}
     \centering
     \includegraphics[width=\textwidth]{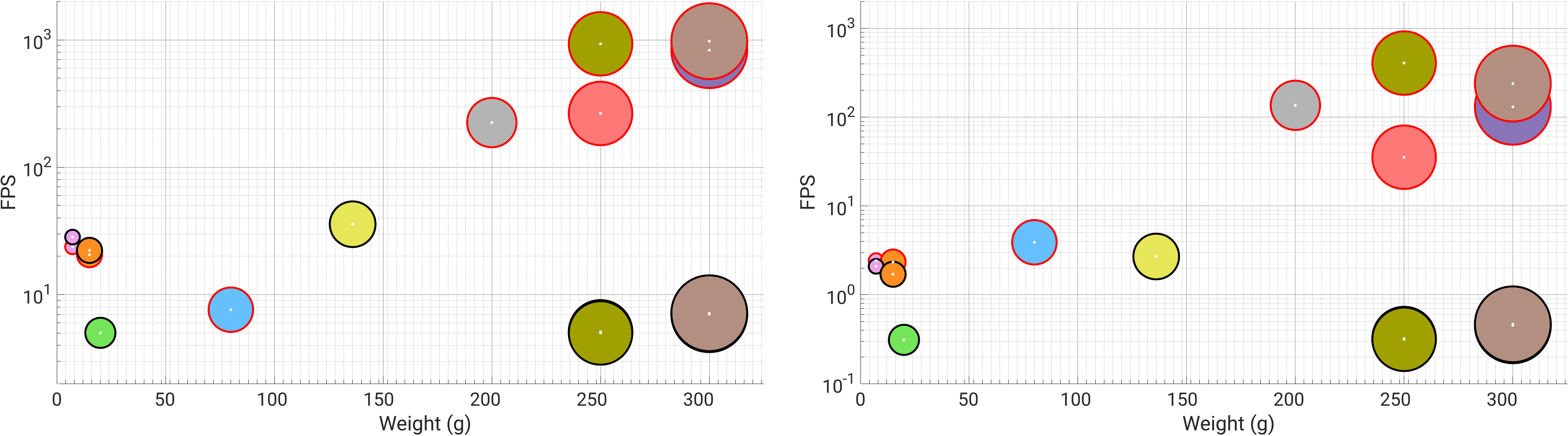}
     \caption{Weight vs. FPS for SqueezeNet$_4$ (T$\times$2, S$\times$2) on different hardware and software optimization combinations. Left: small ($\le$0.83 MB) model, right: large ($\le$8.3 MB) model. The radius of each circle is proportional to log of volume of each hardware.}
     \label{fig:SqueezeNet}
 \end{figure*}

   \begin{figure*}
     \centering
     \includegraphics[width=\textwidth]{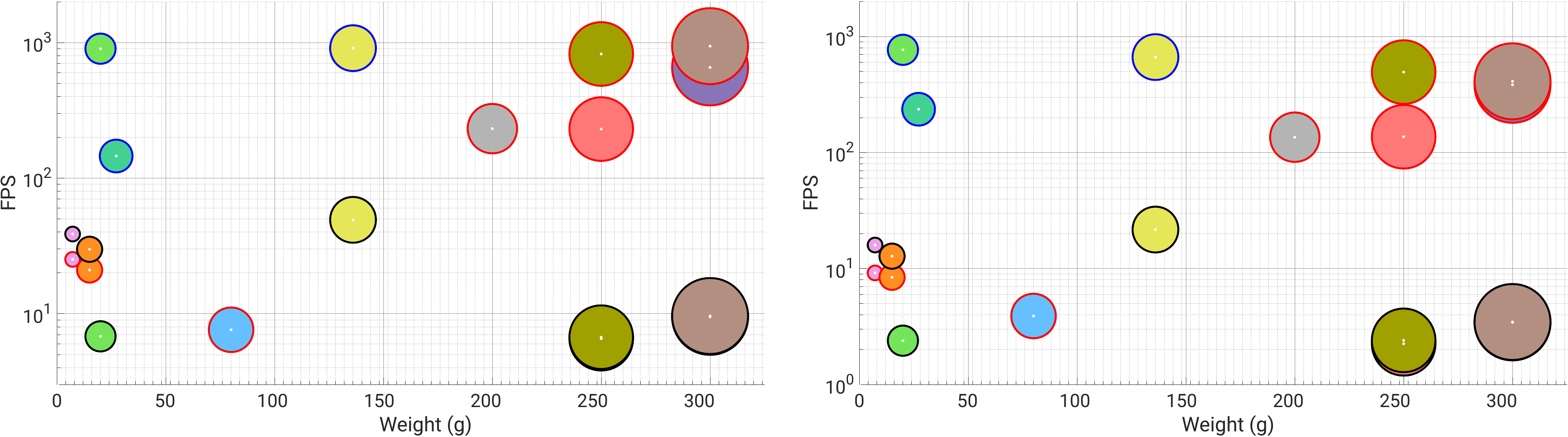}
     \caption{Weight vs. FPS for MobileNet$_4$ (T$\times$2, S$\times$2) on different hardware and software optimization combinations. Left: small ($\le$0.83 MB) model, right: large ($\le$8.3 MB) model. The radius of each circle is proportional to log of volume of each hardware.}
     \label{fig:MobileNet}
 \end{figure*}

   \begin{figure*}
     \centering
     \includegraphics[width=\textwidth]{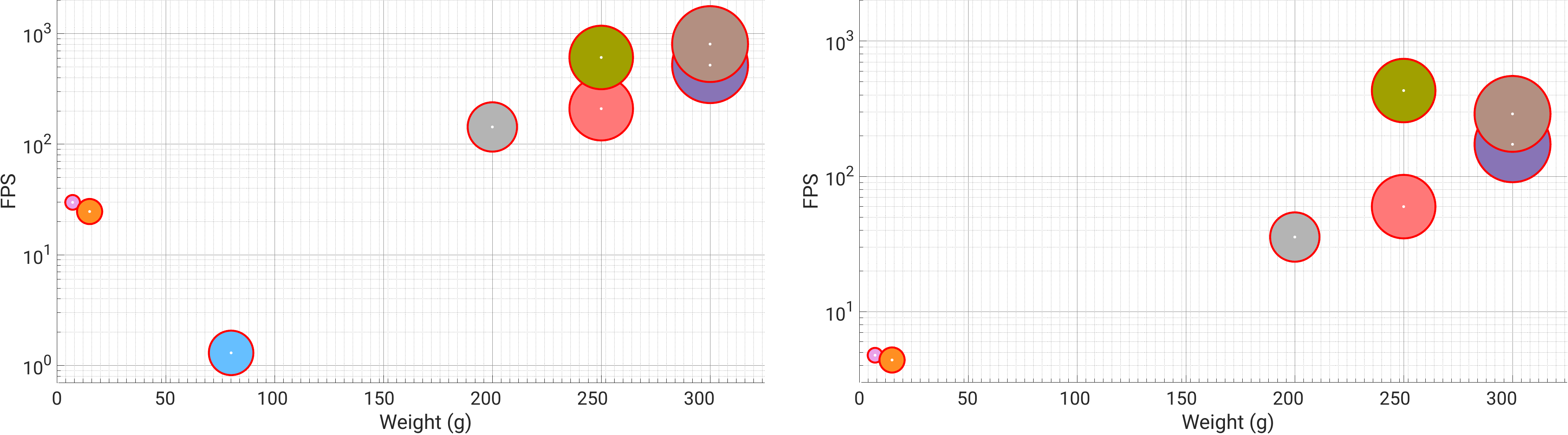}
     \caption{Weight vs. FPS for ShuffleNet$_4$ (T$\times$2, S$\times$2) on different hardware and software optimization combinations. Left: small ($\le$0.83 MB) model, right: large ($\le$8.3 MB) model. The radius of each circle is proportional to log of volume of each hardware.}
     \label{fig:ShuffleNet}
 \end{figure*}

   \begin{figure}
     \centering
     \includegraphics[width=\columnwidth]{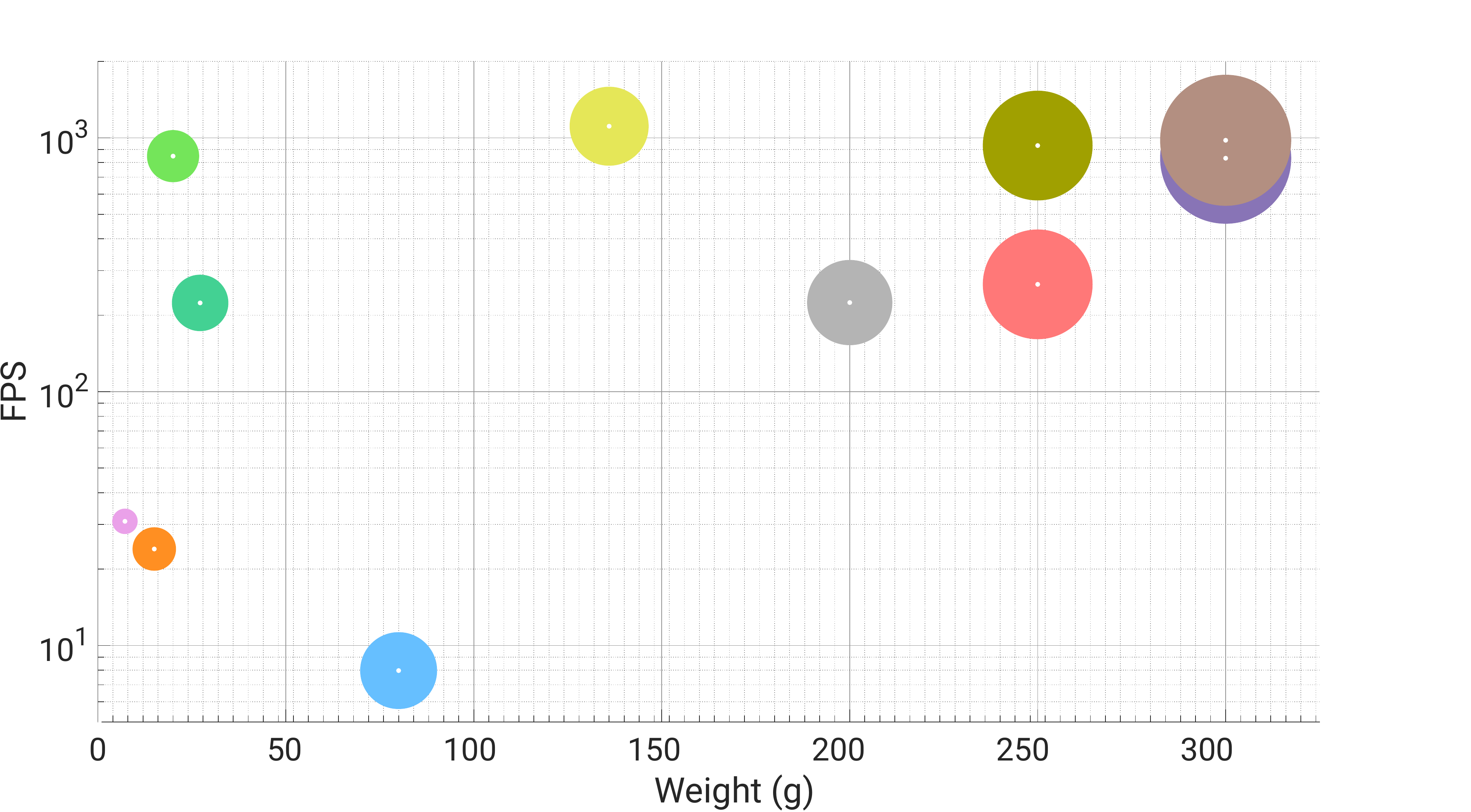}
     \caption{Weight vs. FPS for the best model architecture on each hardware coupled to the best software optimization combination. The radius of each circle is proportional to log of volume of each hardware. The best model architecture and model optimization for each hardware are: Up: ResNet$_S$-\texttt{Float32}, CoralDev: ResNet$_S$-\texttt{Int8-EdgeTPU}, CoralUSB: ResNet$_S$-\texttt{Int8-EdgeTPU}, NanoPi: ResNet$_S$-\texttt{Int8}, BananaPiM2-Zero: ResNet$_S$-\texttt{Int8}, TX2: SqueezeNet$_S$-\texttt{Float32}, Laptop-i7: SqueezeNet$_S$-\texttt{Float32}, Laptop-1070: SqueezeNet$_S$-\texttt{Float32}, PC-i9: SqueezeNet$_S$-\texttt{Float32}, PC-TitanXp: SqueezeNet$_S$-\texttt{Float32}. All networks use T$\times$2, S$\times$2 configuration and $S$ and $L$ subscripts indicate small and large networks respectively. The best network for each hardware was chosen with the avg. error $\le$ 2.5 px. and the configuration which gives the higest FPS.}
     \label{fig:CompileDiffBoard}
 \end{figure}



   \begin{figure}
     \centering
     \includegraphics[width=\columnwidth]{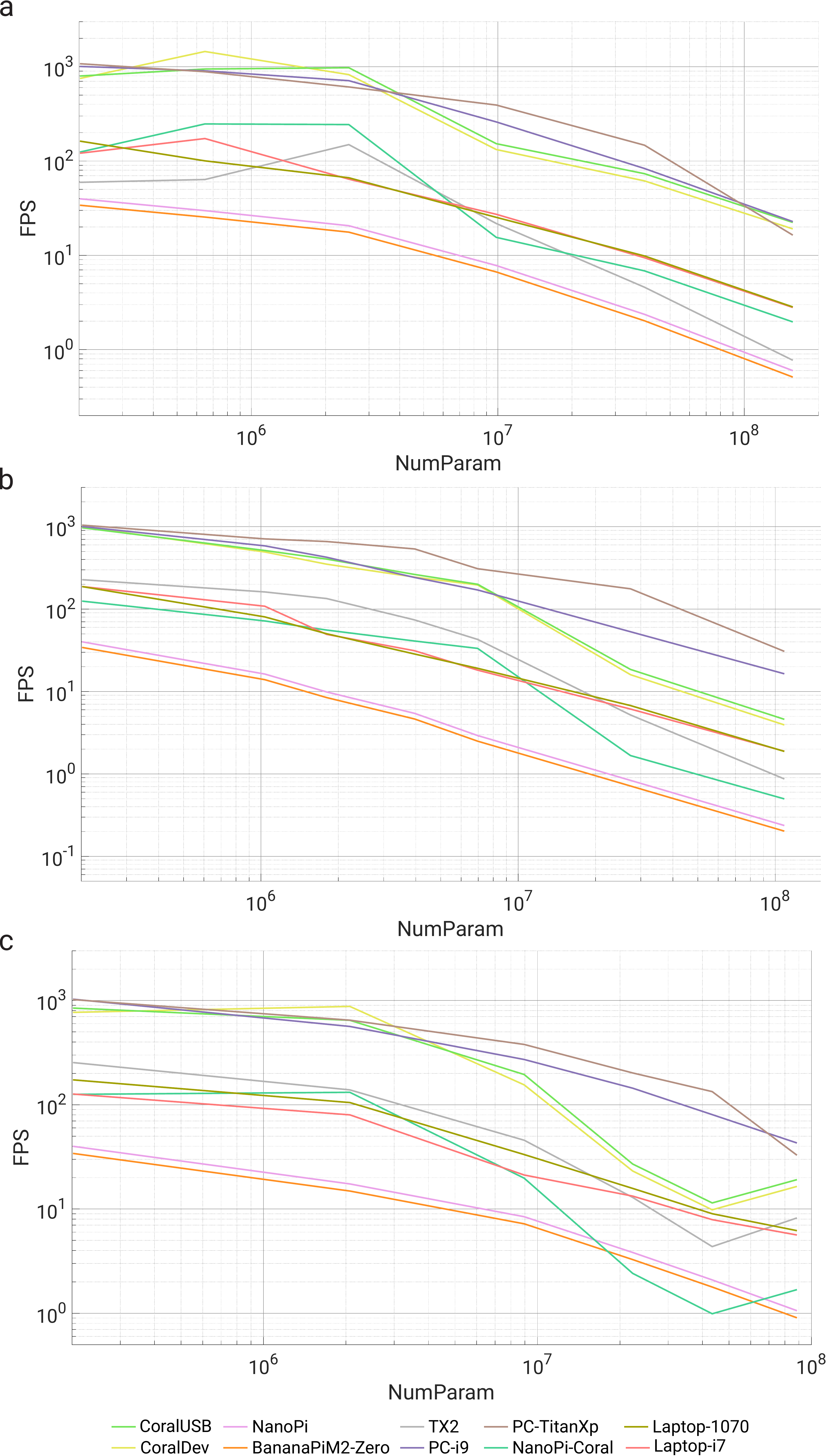}
     \caption{Num. of Params vs. FPS (a) when only increasing the depth of the network while keeping width constant, (b) when only increasing the width of the network while keeping depth constant, (c) when increasing a combination of depth and width of the network for different computers.}
     \label{fig:DiffDeepAndWide}
 \end{figure}

   \begin{figure}
     \centering
     \includegraphics[width=\columnwidth]{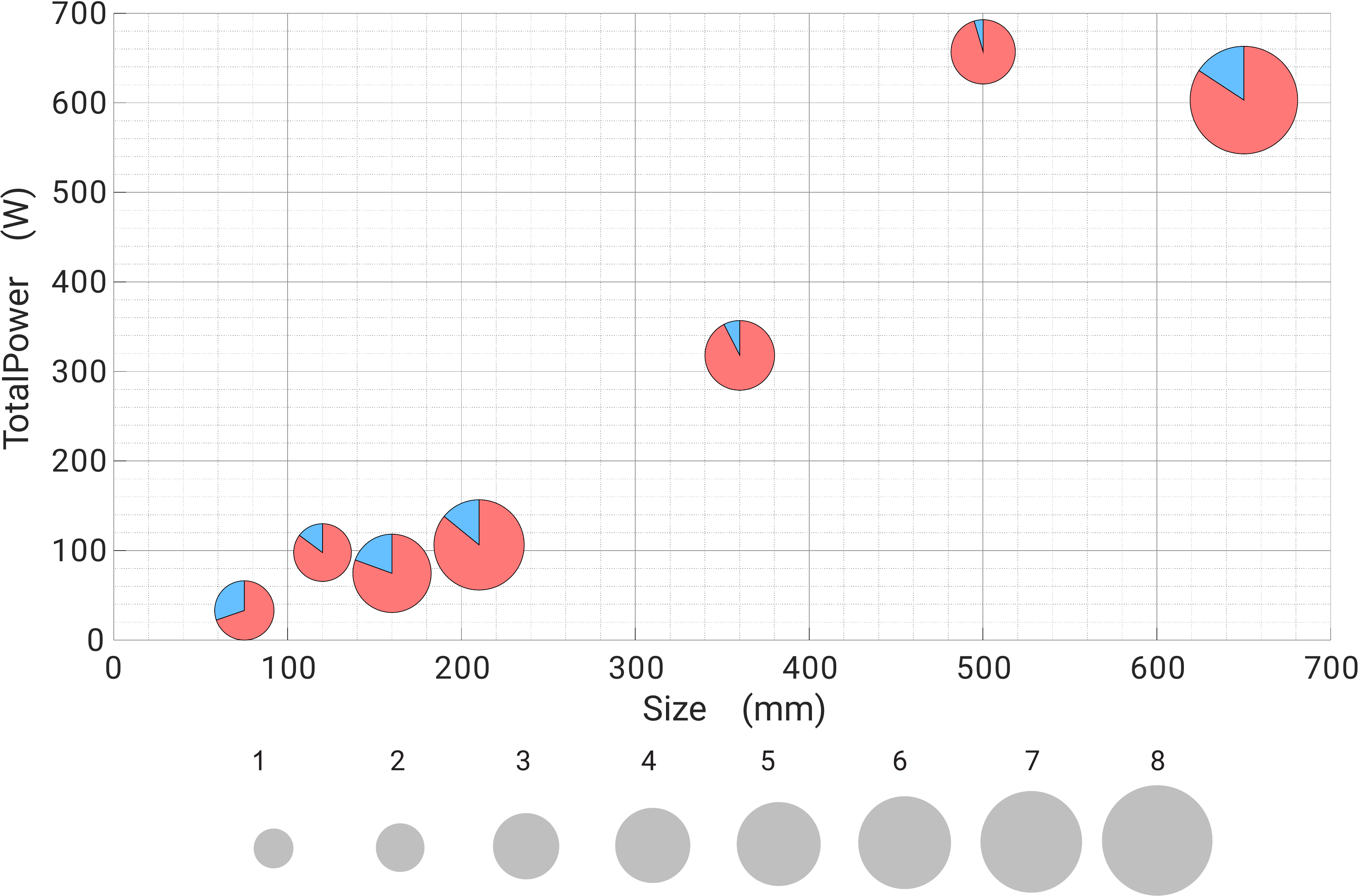}
     \caption{Total Power vs. Quadrotor Size at hover. Each sample is a pie chart which shows the percentage of power consumed by the motors in red and compute and sensing power in blue. The radius of the pie chart is proportional to the power efficiency (in g/W and is given as the ratio of hover thrust to hover power). Refer to the legend on the bottom (gray circles) with the numbers on top indicating power efficiency in g/W.}
     \label{fig:Power}
 \end{figure}

    \begin{figure*}
     \centering
     \includegraphics[width=\textwidth]{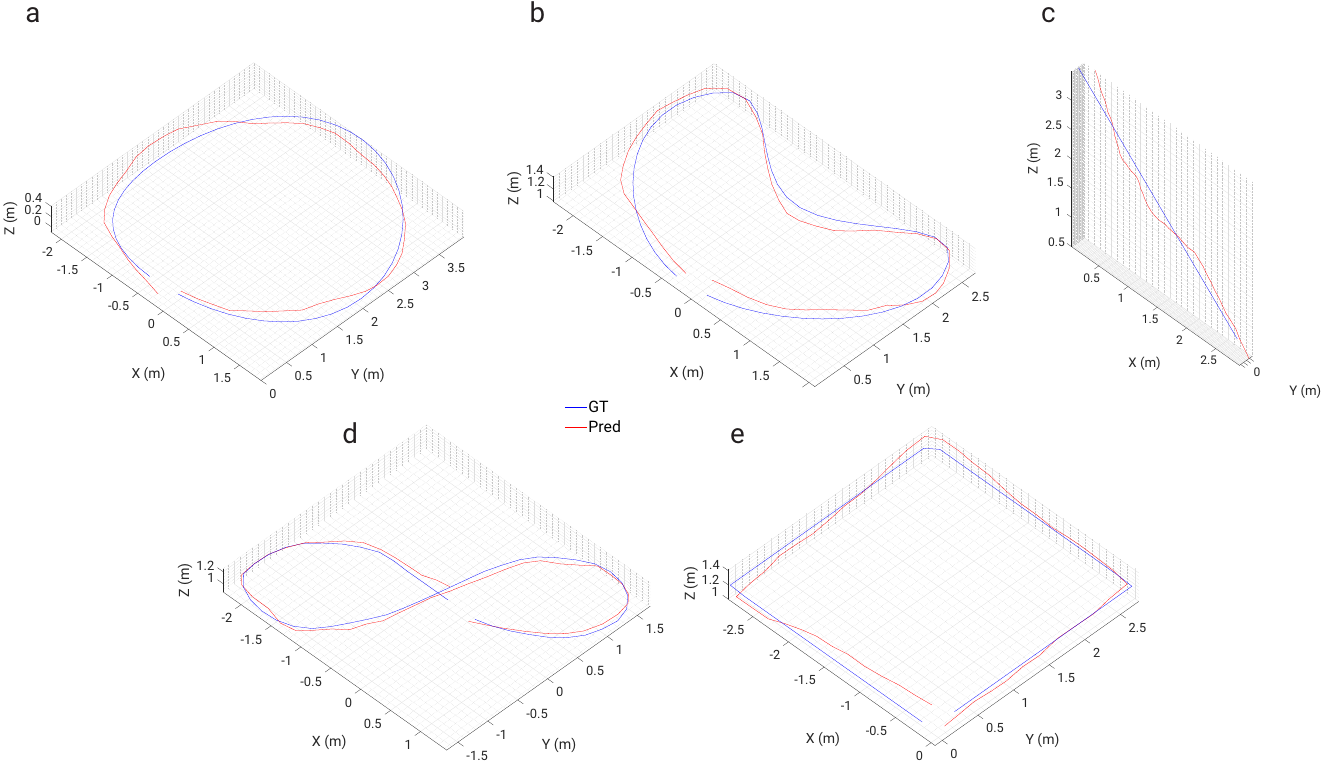}
     \caption{Comparison of trajectory obtained by dead-reckoning (red) our estimates with respect to the ground truth (blue) for quadrotor flight in various trajectory shapes. (a) \texttt{Circle}, (b) \texttt{Moon}, (c) \texttt{Line}, (d) \texttt{Figure8} and (e) \texttt{Square}. }
     \label{fig:TrajPlots}
 \end{figure*}

\subsection{Hardware Aware Design}
We answer the following questions in this section:

\begin{enumerate}
    \item How fast do different network architectures run on a variety of computing platforms subject to weight and volume constraints?
    \item What is the most efficient network architecture for a given hardware abiding the SWAP constraints?
    \item How does varying network width and depth affect the speed of different network architectures on different computing platforms?
    \item Which hardware setup is more power efficient?
    \item How significant is power used for computing compared to power used by other quadrotor components?
\end{enumerate}

\begin{table*}[h!]
\centering
\caption{Different-sized Quadrotor Configuration with respective computers.}
\resizebox{\textwidth}{!}{
\label{tab:QuadConfig}
\begin{tabular}{llllllllll}
\toprule
Quadrotor & Propeller  & \multirow{2}{*}{Motor} & Computing & Computer \multirow{2}{*}{\,\,$\downarrow$} & Total \multirow{2}{*}{$\qquad\,\,\downarrow$} & Auto. Thrust \multirow{2}{*}{$\uparrow$} & Auto. Hover \multirow{2}{*}{$\downarrow$} & Hover \multirow{2}{*}{$\qquad\,\,\downarrow$} & \multirow{2}{*}{$\rho$} \multirow{2}{*}{$\downarrow$} \\
Size (mm) & Size (mm) & & Board & Weight (g) & Weight (g) &  to Weight & Power (W) & Power (W) & \\
\hline

75 & 40 & Happymodel SE0706 15000KV & Nano Pi & 7 & 62 & 1.43 & 33 & 17 & 1.92 \\
120 & 63 & T-Motor F15 Pro KV6000 & Nano Pi + Coral USB & 29 & 209 & 4.67 & 98 & 72 & 1.36 \\
160 & 76 & T-Motor F20II KV3750 & Nano Pi + Coral USB & 29 & 279 & 4.93 & 75 & 49 & 1.53 \\
210 & 152 & EMAX RS2306 KV2750 & Coral Dev Board & 136 & 536 & \textbf{9.91} & 106 & 64 & 1.67 \\
360 & 178 & T-Motor F80 Pro KV1900 & NVIDIA$^\text{\textregistered}$ NVIDIA$^\text{\textregistered}$ Jetson$^\text{TM}$ TX2 & 200 & 1100 & 7.69 & 318 & 222 & 1.43 \\
500 & 254 & iFlight XING 2814 1100KV & NVIDIA$^\text{\textregistered}$ NVIDIA$^\text{\textregistered}$ Jetson$^\text{TM}$ Xavier AGX & 600 & 2000 & 4.98 & 657 & 551 & \textbf{1.19} \\
650 & 381 & Tarot 4414 KV320 & Intel$^\text{\textregistered}$ NUC + NVIDIA$^\text{\textregistered}$ Jetson$^\text{TM}$ Xavier AGX & 1300 & 3900 & 2.72 & 603 & 405 & 1.49 \\
\bottomrule
\end{tabular}}
\end{table*}

\begin{table}[h!]
\centering
\caption{Trajectory evaluation for flight experiemtnts of PRGFlow.}
\resizebox{0.7\columnwidth}{!}{
\label{tab:TrajEval}
\begin{tabular}{llllll}
\toprule
\multirow{2}{*}{Trajectory} & \multicolumn{4}{c}{Error (m) $\downarrow$} & Traj. \\
& X & Y & Z & RMSE & Length (m)\\
\hline
\texttt{Circle} & 0.04 & 0.06 & 0.14 & 0.12 & 12.21 \\ 
\texttt{Moon} & 0.06 & 0.08 & 0.08 & 0.09 & 11.67 \\
\texttt{Line} & 0.06 & 0.06 & 0.10 & 0.09 & 3.84 \\ 
\texttt{Figure8} & 0.03 & 0.03 & 0.05 & 0.05 & 10.91 \\ 
\texttt{Square} & 0.04 & 0.05 & 0.10 & 0.08 & 10.77\\ 
\bottomrule
\end{tabular}}
\end{table}

In the wise words of Alan Kay, a pioneer in computer engineering \textit{``People who are really serious about software should make their own hardware''} one would ideally want to design a hardware chip for a specific SWAP constraint dictated by the size and amount of  features desired in the aerial robot. This can generally only be achieved by the elite drone manufacturing companies owing to the exorbitant non-recurring engineering cost, thereby, putting research labs at a handicap. However, due to the rising Internet of Things (IoT) technologies and computers required to fit tight SWAP constraints researchers can repurpose these computers for efficient utilization on quadrotors (or aerial robots in general). In this spirit, we limit our analysis to commonly used computers designed for IoT purposes which are repurposed for use on aerial robots. The computers used in our experiments are summarized in Table \ref{tab:HardwarePlatforms} and discussed in more detail in Subsec. \ref{subsec:HWPlatforms}.

Refer to Figs. \ref{fig:VanillaNet}, \ref{fig:ResNet}, \ref{fig:SqueezeNet}, \ref{fig:MobileNet} and \ref{fig:ShuffleNet} for a plot of Weight vs. FPS (Frames Per Second) vs. volume of the computer for VanillaNet, ResNet, SqueezeNet, MobileNet and SqueezeNet respectively. All the networks include both small ($\le$0.83 MB) and large ($\le$8.3 MB) configurations training using supervised $l_2$ loss function and optimized using different post-quantization optimizations such as \texttt{Int8-TFLite} and \texttt{Int8-EdgeTPU}. We exclude results from \texttt{Float32-TFLite} due to inferior performance without any significant speedups as compared to the original  \texttt{Float32} model. On can clearly observe that for all the networks the desktop and laptop give the best speed (highest FPS) but also have the largest weights. It would be advisable to gut out a gaming laptop and use it on a larger quadrotor ($\ge$650 mm)  due to the availability of integrated NVIDIA$^\text{\textregistered}$ mobile GPUs with a large amount of CUDA$^\text{\textregistered}$ cores which can be used to accelerate both deep learning and traditional computer vision tasks. An important factor to realize is that using \texttt{Int8-EdgeTPU} optimization to run on either the Coral Dev board or the Coral USB accelerator can provide significant speed-ups of up to 52$\times$ compared to \texttt{Int8-TFLite} without significant loss in accuracy. However, not all operations are supported in TensorFlow Lite and even less operations are supported for EdgeTPU optimization. Also, a drop in speed when going from a smaller model to a larger model is less significant in coral boards due to efficient TPU architecture. Hence, it is advisable to use the Coral Dev board or the Coral USB accelerator whenever possible. We also exclude Intel$^\text{\textregistered}$'s Movidus Neural Compute sticks in our analysis since they provide inferior performance than Coral boards, are harder to use, weigh more and are larger.

A non-obvious observation is that the \texttt{Int8-TFLite} execution speed is much lower than the \texttt{Float32} model on laptops and desktops (both CPU and GPU). This is because of lack of optimized 8-bit integer instruction sets which are generally present in lower-end ARM computers such as the NanoPi and BananaPi. We can also observe a similar drop in performance of up to 2.6$\times$ when AVX and SSE optimized instruction sets are not used on the desktop (Fig. \ref{fig:VanillaNet}, datapoint indicated as PC-i9 NoAVX). The lack of good performance of the Up board can also be pinned to the lack of AVX and SSE instruction sets and should be avoided for neural network tasks if not coupled to a neural network acclerator such as the Intel$^\text{\textregistered}$ Movidius compute stick or the Coral USB accelerator. We also observe speedups of upto 1.7$\times$ on the NanoPi and BananaPi by converting a \texttt{Float32} model to \texttt{Int8-TFLite} due to accelerated NEON instruction sets. Also, note that ShuffleNet models do not support \texttt{Int8-TFLite} and \texttt{Int8-EdgeTPU} optimizations due to unsupported layers. Another interesting observation is that, ResNet (both smaller and larger) and smaller SqueezeNet models achieve almost the same speed on both CPUs and GPUs on the desktop computer.

We choose as the best network architecture and configuration (small versus large) for each hardware, the one which has an average error $\le$2.5 px. and gives the maximum speed. These results are illustrated in Fig. \ref{fig:CompileDiffBoard}. For smaller boards like NanoPi and BananaPi it is recommended to use the smaller ResNet with \texttt{Int8-TFLite} optimization and  for coral boards \texttt{Int8-TFLite} optimization should be used. For the medium sized board TX2 it is advisable to use the smaller SqueezeNet without any optimization (however, we did not explore TensorRT optimization which can result in a 5-20$\times$ speedup since we limit our analysis to the TensorFlow on Python$^\text{TM}$ only environment for ease of use). Again, for larger computers such as a laptop (expect similar performance with a NUC + Jetson Xavier AGX) and a desktop, the smaller SqueezeNet model gives the best performance.

To gather insight into which network dimension (width versus depth) affects the speed the most and to observe the trend on different computers, we measured the speed by varying depth only (Fig. \ref{fig:DiffDeepAndWide}{\color{red}a}), width only (Fig. \ref{fig:DiffDeepAndWide}{\color{red}b}) and width + depth together (Fig. \ref{fig:DiffDeepAndWide}{\color{red}c}). All the networks used in this experiment were VanillaNet$_1$ (PS$\times$1) for consistency. One would expect that increasing depth and width should result in a drop in speed, but this is not always the case. The speeds peak for specific depth/width values (different for different computers) when the perfect balance of memory accesses, size of convolution filters and OPs is achieved. This is more prominent in smaller computers as compared to larger ones (laptop and desktop). This has to be carefully considered when designing networks for use on aerial robots. Also, note that the rate of drop in FPS is more significant with increase in width. A similar trend is observed in Fig. \ref{fig:DiffDeepAndWide}{\color{red}c}. When the depth is increased and the width is decreased there is a sudden increase in FPS towards the end. Finally, performance using Coral with NanoPi can give higher FPS compared to the desktop or laptop when the models are smaller. Hence, using multiple NanoPi + Coral configurations can be more efficient (in-terms of weight and volume) on larger aerial robots as well.

We categorized quadrotors into six standard configurations based on their size -- from 75 mm to 650 mm (pico to large sized), abiding the SWAP constraints. Refer to Table \ref{tab:QuadConfig} and Fig. \ref{fig:Power}. Each quadrotor is configured with a suitable motor, propeller size and most powerful computer that fits the respective quadrotor frame. We define $\rho$ as the ratio of hover power with and without computer. Most literature only talks about the amount of power the on-board computer uses, but this only highlights half of the story. Adding a computer to a quadrotor (to make it autonomous) not only consumes power from the battery but also adds weight which in-turn makes the motor power consumption higher at hover (one would need more thrust to keep the quadrotor flying). Generally, the power efficiency (thrust to power ratio in g/W) decreases with an increase in thrust aggravating the situation further. The essence of this
is quantified by $\rho$ which indicates how much more power  the autonomous quadrotor uses compared to a manual one with the same configuration (excluding computer and sensors). A value of $\rho = 1.0$ is the theoretical best, and the larger the value (above 1.0), the more inefficient the setup. One could clearly observe that the amount of thrust directly increases with quadrotor size as it should but the thrust-to-weight ratio follows a parabolic curve (opening at the top) which achieves a maximum value at 210 mm size. This is due to efficient motor design perfected for racing quadrotors. We also observe that for smaller quadrotors ($\le$75 mm) the power overhead due to adding the computer is significant (as high as 92\%). The power overhead decreases and then increases again as size increases due to the addition of multiple computers at 650 mm size. Also, note that for aggressively flying quadrotors the value of $\rho$ will decrease significantly since we choose the most efficient hovering motors available on the market to maximize battery life. Fig. \ref{fig:DiffQuadrotors} shows four different sized quadrotors, computers and other commonly used quadrotor electronics for a size comparison. We also show how small a hardware designed from the ground-up (Snapdragon flight) can be. We exclude this from our discussions since the smaller model of Snapdragon flight is currently phased out by Qualcomm. Finally, we also experimented with the Sipeed Maix Bit which is a low power neural network accelerator ($<$ 1 W of power) weighing only 20 g with a camera and which can be used on smaller sized quadrotors. However, due to lack of ease of use
we exclude it from our discussion.

\subsection{Trajectory Evaluation}
The predictions $\widetilde{h}$ (VanillaNet$_4$ T$\times$2, S$\times$2 large model trained using supervised $l_2$ loss) are obtained every four frames and are integrated using dead-reckoning to obtain the final trajectory. The trajectory is aligned with the ground truth and evaluated using the approach given in \cite{zhang2018tutorial}. The errors in inidividual axes ($l_1$ distance) and all axes (RMSE) are given for various trajectories in Table \ref{tab:TrajEval} and are illustrated in Fig. \ref{fig:TrajPlots}. We notice that even with simple dead-reakoning we obtain an RMSE of less than 3\% of the trajectory length highlighting the robustness of the proposed PRGFlow.

\section{Summary And Directions For Future Work}
A summary of our observations are given below:
\begin{itemize}
    \item The effect of the choice of network architecture is fairly significant on the accuracy, number of parameters and OPs and needs to be carefully considered when designing a network for deployment on an aerial robot.
    \item Although number of parameters is inversely correlated to speed, sometimes increasing the number of parameters (depth or width) can achieve a speedup due to a better balance in memory accesses, size of convolution filters and OPs which is more significant in smaller computers than larger ones.
    \item It is advisable to use supervised methods for odometry estimation since they are much easier to train and are generally more accurate than their unsupervised counterparts. The loss in accuracy from simulation training to real world is insignificant if enough variation in samples is used which have real world images.
    \item For odometry related regression problems, the simplest method of dropping weights works as well as more complex distillation methods.
    \item Accelerated instruction sets can provide huge speedups and should not be neglected for neural network based methods.
    \item Lastly, deep learning based odometry have two main advantages: they are generally faster due to hardware parallelization and fail more gracefully (work better in adverse scenarios) when compared to their classical counterparts.
    \item A combination of both classical and deep learning methods to solve a problem generally would yield better explainability and preformance gains.
\end{itemize}

Based on the observations and empirical analyses we hope the following directions for future work can further enhance ego-motion/odometry capabilities using deep learning.

\begin{itemize}
 \item Crafting better supervised loss functions (probably on manifolds) should yield more accurate and robust results and probably will surpass advantages of un-supervised/self-supervised methods given enough domain variation.
 \item Formulating problems as nested loops can further simplify problems and provide better explainability than end-to-end deep learning. Also, this enables solving certain problems directly with simple mathematical formulations which can further decrease the number of neurons/OPs required, thereby reducing latency in most cases.
 \item Utilizing multi-frame constraints can significantly boost performance of deep odometry systems.
 \item Better architectures need to be designed to fully utilize the capabilities of un-supervised/self-supervised methods of deep odometry estimation.
\end{itemize}

\section{Conclusions}
We presented a simple way to estimate ego-motion/odometry on an aerial robot using deep learning combining commonly found sensors on-board: a up/down-facing camera, an altimeter source and and IMU. By utilizing simple filtering methods to estimate attitude one can obtain ``cheap'' odometry using attitude compensated frames as the input to a network. We further provided a comprehensive analysis of warping combinations, network architectures and loss functions. All our approaches were benchmarked on different commonly used hardware with different SWAP constraints for speed and accuracy which we hope will serve as a reference manual for researchers and practitioners alike. We also show extensive real-flight odometry results highlighting the robustness of the approach without any fine-tuning or re-training. Finally, as a parting thought, utilizing deep learning when failure is often expected would most likely lead to more robust system.

\section*{Acknowledgement}
The support of the Brin Family Foundation, the Northrop Grumman Mission Systems University Research Program, ONR under grant award N00014-17-1-2622, and the support of the National Science Foundation under grant BCS 1824198 are gratefully acknowledged. We would also like to thank NVIDIA for the grant of a Titan-Xp GPU used in the experiments.

\bibliographystyle{unsrt}
\bibliography{Ref}

\end{document}